\documentclass{article}

\usepackage{a}

\usepackage[utf8]{inputenc} 
\usepackage[T1]{fontenc}    
\usepackage{hyperref}  

\usepackage{hyperref}  
\usepackage{array}
\usepackage{amsmath,amssymb,amsthm,mathtools,bm}
\usepackage{booktabs,enumitem,microtype,needspace} 
\usepackage{lmodern} 
\usepackage[numbers,sort&compress]{natbib}
\newtheorem{theorem}{Theorem}[section]
\newtheorem{lemma}[theorem]{Lemma}

\theoremstyle{definition}

\newcommand{\R}{\mathbb R}
\newcommand{\E}{\mathbb E}
\newcommand{\Prob}{\mathbb P}
\newcommand{\one}{\mathbf 1}

\title{Near-Minimax Sample Complexity for Multi-Objective Reinforcement Learning under Generative Models}
\author{Research draft} 

\author{
  Yue Wang\\
  University of Central Florida
}

\begin{document}
\maketitle

\begin{abstract}
In multi-objective reinforcement learning, preferences may be specified
only after data collection, requiring one dataset to support a continuum
of reward trade-offs. We study finite discounted Markov decision processes
with unknown transitions, $S$ states, $A$ actions, and $m$ known reward
functions taking values in $[0,1]$, given access to a generative model.
The learner must provide an $\varepsilon$-optimal policy for every convex
combination of these rewards without further sampling, with a guarantee
that holds simultaneously over all preferences and initial states.
Writing $H=(1-\gamma)^{-1}$, we establish the constant-confidence minimax
sample complexity
$\widetilde\Theta\!\left(
    \frac{SAH^3}{\varepsilon^2}\min\{S,\log(em)\}
\right),$
where the suppressed factors are logarithmic in $S,A,H,\varepsilon^{-1}$ and the dependence on $m$ is explicit.
Thus, up to these logarithmic factors, the worst-case statistical cost
of supporting all preferences grows logarithmically with the number
of objectives and saturates at the number of states.
The upper bound uses uniform state--action sampling and empirical
planning, with reward rounding in the logarithmic-objective regime.
Its analysis chains optimal-value increments over covers of the reward
hull. A one-sided Bellman variance bound controls signed value
differences under a second transition kernel, preserving the
$H^3\varepsilon^{-2}$ dependence.
A complementary uniform guarantee for all bounded rewards yields the
state-dimension saturation.
We establish lower bounds matching up to logarithmic factors through
a rare-exit construction with subset rewards and action encoding.
These bounds apply to adaptive simulator queries and arbitrary
randomized history-dependent output policies, while the upper bounds
are attained by deterministic stationary policies.
\end{abstract}

\section{Introduction}
Many sequential decision problems involve trade-offs among competing objectives. A robot may balance task completion, energy consumption, and wear, and a language model may be evaluated for both helpfulness and safety. Multi-objective reinforcement learning (MORL) represents such criteria as separate reward signals and studies how the resulting decisions depend on the user's preferences~\cite{roijers2013survey,hayes2022practical}. We consider linear scalarization. The learner knows $m$ rewards $r_1,\ldots,r_m$, which we call the coordinate rewards; together they form the reward dictionary $\{r_1,\ldots,r_m\}$. A preference vector $w\in\Delta_m$ assigns them nonnegative weights summing to one and induces the scalar reward $r_w=\sum_{i=1}^m w_ir_i$. The dynamics are shared across preferences, but the optimal policy can change with the weights.
 
Preferences are often specified only after data collection. Different users may weigh the same objectives differently, and a designer may inspect the available trade-offs before fixing a reward. Preference-conditioned value functions~\cite{yang2019generalized} and interpolation between models fine-tuned on different rewards~\cite{rame2023rewarded} address this need for adaptation. The statistical question is what a single dataset can guarantee. For one fixed scalar reward, the dataset must support one near-optimal policy. For a finite list of rewards, it must support a near-optimal policy for each listed reward. Uniform MORL asks for a policy-selection rule that is simultaneously accurate for every convex combination of the coordinate rewards. This requirement covers a continuum of rewards, including preferences for which no policy that is optimal for a single coordinate reward is optimal.
 
We study this uniform target in finite discounted Markov decision processes (MDPs) with unknown transitions and access to a generative model, that is, a simulator that returns a next state for any queried state--action pair. Write $S$ and $A$ for the numbers of states and actions, $\gamma\in[0,1)$ for the discount factor, and $H=(1-\gamma)^{-1}$ for the effective horizon. For one fixed bounded reward, the minimax sample complexity is $\widetilde\Theta(SAH^3/\varepsilon^2)$ in the standard nontrivial parameter regime~\cite{azar2013minimax,agarwal2020model,li2020breaking}. The closest discounted MORL analysis gives an $\widetilde O(mSAH^3/\varepsilon^2)$ bound for estimating scalarized $Q$-functions uniformly over preferences~\cite{zhou2020provable}. A direct reduction through a net of preferences also gives linear dependence on $m$ for uniform policy learning. For a finite list of known rewards, recent online episodic results instead depend logarithmically on the list size~\cite{chen2026optimal}. These results motivate the following question.
\begin{center}
\fbox{\begin{minipage}{0.92\linewidth}
\centering\itshape
How many simulator queries are needed to support a near-optimal policy for every reward in the convex hull of a known reward dictionary?
\end{minipage}}
\end{center}
 
Our learner collects one dataset and then answers every preference query without further simulator calls. Its guarantee holds on a single high-probability event, simultaneously for all preferences and all initial states (Section~\ref{subsec:simulator}). Specifically, our contributions are summerized as follows.

\textbf{A near-minimax characterization of the objective dependence.}
For $S\ge3$, $A\ge2$, $H\ge2$, and $\varepsilon\le H/4096$, we show that the minimax sample complexity is
\[
\widetilde\Theta\!\left(
    \frac{SAH^3}{\varepsilon^2}
    \min\{S,\log(em)\}
\right),
\qquad H\triangleq(1-\gamma)^{-1},
\]
where the suppressed factors are logarithmic in $S,A,H,\varepsilon^{-1}$ and the dependence on $m$ is retained explicitly (Theorem~\ref{thm:main}). Supporting every convex combination of the known rewards thus costs only a logarithmic factor in the number of objectives, and this factor never exceeds the number of states. The upper bound combines two procedures. Empirical planning for a rounded reward gives the $\log(em)$ factor (Theorem~\ref{thm:chaining}). Direct empirical planning for the requested reward gives the factor $S$, since its guarantee holds for every reward in $[0,1]^{SA}$ (Theorem~\ref{thm:reward-free-simulator}). The learner uses whichever procedure needs fewer samples. Each procedure samples every state--action pair equally once, and returns deterministic stationary policies without further simulator queries.

\textbf{A chaining analysis for uniform policy learning.}
Nearby preferences have nearby optimal values, so the data needed to serve them overlap; the difficulty is to exploit this overlap without losing accuracy. The upper bound rests on finite covers of the convex reward hull. At resolution $\eta$, their logarithmic size is of order $\eta^{-2}\log(SA)\log(em)$. A single union bound over the cover at the final resolution $\varepsilon/H$ would add a factor of order $H^2/\varepsilon^2$. We instead write each optimal value as a sum of increments across successively finer reward approximations, and the variance of each increment shrinks with its resolution. The central technical step is a one-sided Bellman variance bound for signed differences of optimal values, with variances taken under a second transition kernel (Lemmas~\ref{lem:resolvent-variance} and~\ref{lem:chained-comparison}). The bound uses only an upper bound on the one-step drift. It therefore applies even though the policy that is optimal for the target reward need not be optimal for its approximants. Increments of empirical optimal values depend on the data used to compute them. An adaptation of the absorbing-state construction of~\cite{agarwal2020model} supplies concentration for these increments. The variance bound then turns the row-wise concentration into bounds on policy values, in both directions of the comparison between the true and empirical models. Across levels, the growth of the cover sizes is balanced by the decay of the increment variances, which preserves both the $H^3$ and the $\varepsilon^{-2}$ dependence.

\textbf{Lower bounds with bounded rewards and simplex preferences.} We construct stationary discounted MDPs in which answering every preference query accurately identifies many unknown exit distributions. Rare exits produce the $H^3/\varepsilon^2$ dependence. Subset rewards distinguish distributions over $d$ shared absorbing outcomes using $2^d$ coordinate rewards, which yields a factor of order $\log m$, capped by $S$. An additional action code of length $O(\log(A+1))$ lets a simplex preference single out one pair of actions while keeping every reward in $[0,1]$ (Theorem~\ref{thm:simulator-lower-action-code}). The lower bounds allow adaptive simulator queries and randomized history-dependent output policies. For any fixed output policy, its normalized pre-exit action weights do not depend on the unknown exit distributions. This identity lets the learned policy-selection rule serve as a decoder. A scalar construction covers the remaining parameter regimes (Theorem~\ref{thm:simulator-lower}).

\subsection{Related work}\label{subsec:related}
 
\textbf{MORL targets and Pareto geometry.}
Surveys distinguish MORL problems by their utility functions, policy classes, and the information available about preferences~\cite{roijers2013survey,hayes2022practical}. For linear utilities, a convex coverage set contains an optimal policy for every preference, whereas Pareto-front recovery describes the attainable nondominated value vectors. Lu et al.~\cite{lu2023multi} study convexity and the relations among Pareto notions for stationary policies. Li et al.~\cite{li2024how} exploit geometric locality to recover the exact Pareto front in a known model. Other learning targets include approachability of a target set in average-reward problems~\cite{mannor2004geometric} and convergence to Pareto-stationary policies~\cite{zhou2024finite}. Our target is uniform scalarized policy accuracy in an unknown model, delivered by a policy-selection rule. Section~\ref{subsec:simulator} explains how this rule acts as an implicit approximate convex coverage set.
 
\textbf{Preference-conditioned and offline methods.}
Abels et al.~\cite{abels2019dynamic} develop deep MORL methods for changing weights, and Yang et al.~\cite{yang2019generalized} use an envelope Bellman update to share one value representation across preferences. Zhu et al.~\cite{zhu2023scaling} learn preference-adaptive policies from offline demonstrations, and Ram\'e et al.~\cite{rame2023rewarded} interpolate models fine-tuned on different rewards. These works study representations and algorithms for preference adaptation. We instead quantify the worst-case number of transition samples needed for a simultaneous policy guarantee in a tabular discounted model.
 
\textbf{Generative-model sample complexity.}
For a single reward, variance-sensitive arguments and empirical planning give sharp sample bounds~\cite{azar2013minimax,agarwal2020model,li2020breaking}. Zhou et al.~\cite{zhou2020provable} extend model-based envelope value iteration to MORL and obtain a bound linear in $m$ for a preference-weighted $Q$-function error. This value-estimation target differs from the policy loss in the true model studied here. A preference net combined with scalar empirical-planning guarantees yields a uniform policy bound with the same linear dependence on $m$. Our analysis instead uses multiscale covers of the reward hull to obtain the logarithmic factor. In a related simulator setting, Vaswani et al.~\cite{vaswani2022near} characterize the cost of relaxed and strict feasibility in constrained MDPs. Their objective is a constrained scalar optimum, whereas ours is a family of unconstrained scalar rewards.
 
\textbf{Online multi-objective learning.}
Wu et al.~\cite{wu2021accommodating} study adversarial preferences and preference-free exploration in finite-horizon MDPs. With $H_{\mathrm{ep}}$ denoting the episode length, their exploration upper bound is $\widetilde O(\min\{m,S\}SAH_{\mathrm{ep}}^3/\varepsilon^2)$ episodes. Qiu et al.~\cite{qiu2024traversing} develop online and preference-free algorithms for Tchebycheff and smooth Tchebycheff scalarizations. Hu and Yu~\cite{hu2026near} study near-minimax learning under predictable adversarial preferences and preference-free exploration in linear MDPs. These works differ from ours in the interaction protocol, the scalarization, or the model class. We study linear scalarization in a stationary discounted tabular model with direct state--action sampling.
 
\textbf{Finite reward lists and reward-free exploration.}
Reward-free exploration collects data before any reward is specified and must support arbitrary bounded rewards afterward~\cite{jin2020reward}. Task-agnostic exploration instead considers a finite collection of tasks whose rewards are unavailable during exploration~\cite{zhang2020task}. Later work sharpens these guarantees~\cite{menard2021fast,li2026minimax,ridel2026improved}. In particular, Li et al.~\cite{li2026minimax} separate reward-agnostic learning for a finite reward class from the all-rewards requirement. Ridel and Cohen~\cite{ridel2026improved} reduce the lower-order term that restricts the accuracy range of the reward-agnostic bound. They also prove a reward-free lower bound of order $S^2AH_{\mathrm{ep}}^3/\varepsilon^2$ for time-inhomogeneous MDPs, which matches the reward-free upper bound up to logarithmic factors. For rewards known before exploration, Chen and Zhang~\cite{chen2026optimal} give an online episodic upper bound of $\widetilde O(SAH_{\mathrm{ep}}^3\log m/\varepsilon^2)$ episodes for a list of $m$ rewards. Under the parameter conditions of their Corollary~1, they also prove a lower bound of order $SAH_{\mathrm{ep}}^3\min\{S,\log m\}/\varepsilon^2$. Our upper bound covers the entire convex hull of the given rewards, not only the list. Our lower bounds use a stationary discounted kernel and count individual simulator observations. Episodic bounds count trajectories, so their horizon factors cannot be compared directly with $H=(1-\gamma)^{-1}$. Theorem~\ref{thm:reward-free-simulator} supplies the complementary all-rewards guarantee under our sampling model. Table~\ref{tab:prior} collects these rates.
 
\begin{table}[t]
\centering
\footnotesize
\newcommand{\He}{H_{\mathrm{ep}}}
\renewcommand{\arraystretch}{1.25}
\setlength{\tabcolsep}{4pt}
\begin{tabular}{@{}>{\raggedright\arraybackslash}p{0.16\textwidth}>{\raggedright\arraybackslash}p{0.12\textwidth}>{\raggedright\arraybackslash}p{0.245\textwidth}>{\raggedright\arraybackslash}p{0.205\textwidth}>{\raggedright\arraybackslash}p{0.20\textwidth}@{}}
\toprule
Work & Access & Rewards served & Upper bound $\times\varepsilon^2$ & Lower bound $\times\varepsilon^2$\\
\midrule
Zhou et al.~\cite{zhou2020provable} & generative, discounted & $Q$-values for all $w$ with $\|w\|_1\le1$ & $mSAH^3$ & single objective only\\
Net + planning~\cite{agarwal2020model,li2020breaking} & generative, discounted & $\Delta_m$ & $mSAH^3$ & --\\
Wu et al.~\cite{wu2021accommodating} & online, episodic & $\Delta_m$, preferences revealed later & $\min\{m,S\}SA\He^3$ & $\min\{m,S\}SA\He^2$\,$^{\dagger}$\\
Jin et al.~\cite{jin2020reward} & online, episodic & all rewards, revealed later & $S^2A\He^5$ & $S^2A\He^2$\\
Zhang et al.~\cite{zhang2020task} & online, episodic & $m$ tasks, unknown during exploration & $SA\He^5\log m$ & $SA\He^2\log m$\\
Li et al.~\cite{li2026minimax} & online, episodic & $\mathrm{poly}(S,A,\He)$ given rewards; all rewards & $SA\He^3$\,$^{\ddagger}$; $S^2A\He^3$ & --\\
Ridel and Cohen~\cite{ridel2026improved} & online, episodic & $\mathrm{poly}(S,A,\He)$ given rewards; all rewards & $SA\He^3$; $S^2A\He^3$ & --; $S^2A\He^3$\,$^{\S}$\\
Chen and Zhang~\cite{chen2026optimal} & online, episodic & $m$ known rewards & $SA\He^3\log m$ & $SA\He^3\min\{S,\log m\}$\,$^{\S}$\\
\midrule
This work & generative, discounted & $\mathcal R$ (equivalently $\Delta_m$) & $SAH^3\min\{S,\log(em)\}$ & $SAH^3\min\{S,\log(em)\}$\\
\bottomrule
\end{tabular}
\caption{Sample-complexity results for learning near-optimal policies for many rewards in tabular MDPs. Each entry is the leading term of the bound multiplied by $\varepsilon^2$; constants and logarithmic factors other than those shown are omitted, and $m$ is the number of rewards or objectives. Online episodic rows count episodes of length $H_{\mathrm{ep}}$ with per-step rewards in $[0,1]$. Generative rows count simulator queries in discounted MDPs with $H=(1-\gamma)^{-1}$. Rows with different access models use different units and are not directly comparable. The row ``Net + planning'' combines a preference net with the cited scalar guarantees. Linear-MDP results~\cite{hu2026near} and non-linear scalarizations~\cite{qiu2024traversing} are not listed.
$^{\dagger}$Stated in~\cite[Theorem~4]{wu2021accommodating}. Its proof uses~\cite[Appendix~D, Lemma~12]{wu2021accommodating}, which is stated for preferences on the unit $\ell_1$ sphere; under that normalization, all actions in its construction are within $2\varepsilon/m$ of each other in scalarized value, so the lemma does not hold as stated.
$^{\ddagger}$A lower-order term of order $S^4A^4\He^6/\varepsilon$ dominates unless $\varepsilon$ is small; \cite{ridel2026improved} reduces it to $S^2A\He^4/\varepsilon$.
$^{\S}$Time-inhomogeneous constructions; the bound of~\cite{chen2026optimal} holds under the parameter conditions of its Corollary~1.}
\label{tab:prior}
\end{table}
 
\section{Problem formulation}\label{sec:formulation}
 
\subsection{Model and reward class}\label{subsec:model}
Let $\mathcal S$ and $\mathcal A$ be finite sets with cardinalities $S$ and $A$. The stationary transition kernel $P(\cdot\mid s,a)$ is unknown. The discount factor $\gamma\in[0,1)$ is known, and $H\triangleq(1-\gamma)^{-1}$. The learner is given the complete tables of the deterministic coordinate rewards $r_i:\mathcal S\times\mathcal A\to[0,1]$, $i\in[m]$, which form the reward dictionary $\{r_1,\ldots,r_m\}$. We identify each reward with its table in $[0,1]^{SA}$. Define
\[
 \Delta_m\triangleq\Bigl\{w\in\mathbb R^m:\ w_i\ge0,\ \sum_{i=1}^m w_i=1\Bigr\},\qquad
 r_w\triangleq\sum_{i=1}^m w_ir_i,\qquad
 \mathcal R\triangleq\operatorname{conv}\{r_1,\ldots,r_m\}.
\]
We call $w\in\Delta_m$ a preference and $\mathcal R$ the reward hull. Since $\mathcal R=\{r_w:w\in\Delta_m\}$, every scalarized reward lies in $[0,1]^{SA}$, whatever the value of $m$.
 
A policy $\pi$ may be randomized and history-dependent. For a reward $r\in[0,1]^{SA}$, we use unnormalized discounted values,
\[
 V_r^\pi(s)\triangleq\mathbb E_s^{P,\pi}\Bigl[\sum_{t=0}^\infty\gamma^tr(s_t,a_t)\Bigr],\qquad
 V_r^*(s)\triangleq\sup_\pi V_r^\pi(s),
\]
where $\mathbb E_s^{P,\pi}$ denotes expectation over trajectories with $s_0=s$, actions drawn from $\pi$, and transitions drawn from $P$. Both values lie in $[0,H]$. For a preference $w$, we abbreviate $V_w^\pi\triangleq V_{r_w}^\pi$ and $V_w^*\triangleq V_{r_w}^*$. Every guarantee below holds at each initial state, not only on average over a prescribed initial distribution.
 
\subsection{Which rewards must one dataset support?}\label{subsec:targets}
We further distinguish different learning targets, determined by two features: the reward information available during sampling, and the query class, that is, the set of rewards for which the learner must later return a near-optimal policy. Table~\ref{tab:targets} compares four targets under a common discounted simulator model.
 
\textbf{A fixed reward versus a finite list.}
If $w_0$ is fixed and known before sampling, optimizing $r_{w_0}$ is an ordinary single-reward learning problem, and the sampling procedure may be tailored to that reward. For the finite list $\{r_1,\ldots,r_m\}$, one dataset must support policies $\widehat\pi_1,\ldots,\widehat\pi_m$ that are simultaneously near-optimal for their respective rewards. This joint guarantee on the listed rewards says nothing about other preferences.
 
\textbf{A finite list versus its convex hull.}
Uniform MORL also covers every mixed preference, and policies that are optimal for individual coordinate rewards need not suffice. Consider a single-state problem with three actions whose reward vectors $(r_1(s,a),r_2(s,a))$ are $(1,0)$, $(0,1)$, and $(0.6,0.6)$. The first two actions are optimal for $r_1$ and $r_2$, respectively. The third is optimal for $w=(1/2,1/2)$, and no randomization between the first two attains its value. This distinction concerns the required outputs. By itself, it does not imply a larger minimax sample complexity for the hull than for the finite list.
 
\begin{table}[t]
\centering
\small
\renewcommand{\arraystretch}{1.2}
\begin{tabular}{@{}l>{\raggedright\arraybackslash}p{0.37\textwidth}>{\raggedright\arraybackslash}p{0.29\textwidth}@{}}
\toprule
Learning target & Reward information during sampling & Query class\\
\midrule
Fixed reward & One scalar reward $r_{w_0}$ is known & The singleton $\{r_{w_0}\}$\\
Finite known tasks & The rewards $r_1,\ldots,r_m$ are known & The finite set $\{r_1,\ldots,r_m\}$\\
Uniform MORL (this work) & The rewards $r_1,\ldots,r_m$ are known; the query preference is not & The reward hull $\mathcal R$\\
Reward-free learning & No reward is known & Every reward in $[0,1]^{SA}$\\
\bottomrule
\end{tabular}
\caption{Learning targets, distinguished by the reward information available during sampling and by the query class. The last three rows require simultaneous accuracy over the whole query class from one dataset. All rows assume the discounted simulator model of Section~\ref{subsec:simulator}; Table~\ref{tab:prior} also lists results for trajectory-based interaction.}
\label{tab:targets}
\end{table}
 
\textbf{A known reward hull versus arbitrary rewards.}
In our setting, the reward dictionary is available before sampling, although no query preference is specified. Reward-free learning must instead cover the entire bounded reward class, without this prior restriction~\cite{jin2020reward}. The inclusions
\[
 \{r_1,\ldots,r_m\}\subseteq\mathcal R\subseteq[0,1]^{SA}
\]
order the query classes of finite-task learning, uniform MORL, and reward-free learning. Reward knowledge during sampling and the requirement to serve every query afterward are separate features of a target. Our upper-bound procedures do not use the reward values during sampling: they sample every state--action pair equally often. The known dictionary enters through the sample size, which depends on $m$, and through the policy selection after sampling.
 
\subsection{Simulator access and the simultaneous guarantee}\label{subsec:simulator}
Each generative-model query selects a state--action pair $(s,a)$ and returns a fresh draw from $P(\cdot\mid s,a)$, conditionally independent of earlier observations given the queried pair. Query choices may depend on the preceding transcript. The sample complexity counts these next-state observations, so we use ``sample'' and ``simulator query'' interchangeably. The reward tables are known and require no sampling.
 
After training, the learner returns a policy-selection rule $w\mapsto\widehat\pi_w$. Answering a query may use the stored data and additional computation, but makes no further simulator calls. Let $\mathcal D$ denote the training transcript together with the learner's internal random seed, so that the selection rule is fixed once $\mathcal D$ is fixed. Define the loss of the rule at preference $w$ by
\[
 \mathcal E_{\mathcal D}(w)\triangleq\max_{s\in\mathcal S}\bigl(V_w^*(s)-V_w^{\widehat\pi_w}(s)\bigr).
\]
For every transition kernel and every reward dictionary, we require the event
\begin{equation}\label{eq:learning-target}
 \sup_{w\in\Delta_m}\mathcal E_{\mathcal D}(w)
 =\sup_{w\in\Delta_m}\max_{s\in\mathcal S}\bigl(V_w^*(s)-V_w^{\widehat\pi_w}(s)\bigr)\le\varepsilon
\end{equation}
to hold with probability at least $1-\delta$:
\begin{equation}\label{eq:prob-guarantee}
 \mathbb P_{\mathcal D}\Bigl(\sup_{w\in\Delta_m}\mathcal E_{\mathcal D}(w)\le\varepsilon\Bigr)\ge1-\delta.
\end{equation}
The probability is over the training observations and the algorithmic randomness. Randomness during deployment, when present, is already averaged into the policy values. We refer to \eqref{eq:learning-target}--\eqref{eq:prob-guarantee} as the uniform MORL guarantee.
 
A pointwise guarantee for the same sampling procedure would instead require
\[
 \mathbb P_{\mathcal D}\bigl(\mathcal E_{\mathcal D}(w)\le\varepsilon\bigr)\ge1-\delta\qquad\text{for every fixed }w\in\Delta_m.
\]
Its exceptional datasets may depend on $w$. Our requirement places the supremum over $w$ inside one probability event, so a successful dataset supports all preferences at once, including preferences chosen after inspecting the training output. The rule may select different policies for different preferences; no single policy is required to be optimal for every reward.
 
The rule acts as an implicit $\varepsilon$-approximate convex coverage set, with accuracy required at every initial state: rather than listing policies, it returns an $\varepsilon$-optimal policy for any queried preference. Exact enumeration of the Pareto front, or of all policies attaining a Pareto-optimal value, is a different output requirement. The upper bounds return deterministic stationary policies, whereas the lower bounds allow arbitrary randomized history-dependent output policies. The minimax quantity counts simulator queries only; we do not analyze the computational cost of storing or evaluating the policy-selection rule.
 
\subsection{Notation and standard planning facts}\label{subsec:notation}
A transition kernel $K$ assigns a distribution $K(\cdot\mid s,a)$ on $\mathcal S$ to each state--action pair; the two kernels used below are the true kernel $P$ and the empirical kernel $\widehat P$ of Section~\ref{sec:chaining}. For a reward $r$, let $v_r(K)\in\mathbb R^S$ be the optimal value vector in $K$, and let $V_r^\pi(K)$ be the value vector of a policy $\pi$ in $K$. An omitted kernel is $P$, so $v_r=v_r(P)=V_r^*$ and $V_r^\pi=V_r^\pi(P)$. For a deterministic stationary policy $\pi$, write $r_\pi(s)\triangleq r(s,\pi(s))$, and let $K_\pi$ be the $S\times S$ stochastic matrix whose row $s$ is $K(\cdot\mid s,\pi(s))$. For $f\in\mathbb R^S$, write
\[
 Kf(s,a)\triangleq\sum_{s'\in\mathcal S}K(s'\mid s,a)f(s'),\qquad
 \sigma_K(f)(s,a)\triangleq\sqrt{K(f^2)(s,a)-\bigl(Kf(s,a)\bigr)^2}.
\]
For $K_\pi$, the same notation is indexed by states. Thus $K_\pi f(s)=Kf(s,\pi(s))$, and similarly $\sigma_{K_\pi}(f)(s)=\sigma_K(f)(s,\pi(s))$. Vector inequalities are coordinatewise. Squares, absolute values, and products of vectors are also coordinatewise, $\mathbf 1$ is the all-ones vector, and $\|f\|_\infty$ is the largest absolute entry of a vector or array.
 
Finite discounted MDPs admit deterministic stationary policies that are optimal at every state, even when the comparison class contains all randomized history-dependent policies~\cite[Theorem~6.2.10]{puterman1994}. For every kernel $K$, every policy $\pi$, and all $r,q\in[0,1]^{SA}$, Bellman contraction and the discounted reward sum give
\begin{equation}\label{eq:lipschitz}
\begin{aligned}
 0\le v_r(K)&\le H\mathbf 1,\\
 \|v_r(K)-v_q(K)\|_\infty&\le H\|r-q\|_\infty,\\
 \|V_r^\pi(K)-V_q^\pi(K)\|_\infty&\le H\|r-q\|_\infty.
\end{aligned}
\end{equation}
In particular, $\|r_w-r_{w'}\|_\infty\le\|w-w'\|_1$ implies that optimal values are $H$-Lipschitz in the preference with respect to $\|\cdot\|_1$, although an optimal policy need not vary continuously with the preference.

\section{Main result: Near Minimax Sample Complexity}\label{sec:main}
Let $N^*(S,A,m,H,\varepsilon)$ be the smallest worst-case maximum number of simulator queries sufficient for~\eqref{eq:learning-target} with success probability at least $3/4$, uniformly over all known reward functions $r_1,\ldots,r_m:\mathcal S\times\mathcal A\to[0,1]$ and all transition kernels. Computational cost is not included in this definition.
Set
\[
\kappa(S,m)\triangleq \min\{S,\log(em)\},\qquad
L\triangleq \log\frac{eSAH^2}{\varepsilon},\qquad
L_A\triangleq 1+\log\log(e^e A).
\]
Natural logarithms are used unless a base is displayed.

\begin{theorem}[Joint near-minimax characterization]
\label{thm:main}
There are universal constants $c,C>0$ such that, for $S\ge3$, $A\ge2$, $m\ge1$, $H\ge2$, and $0<\varepsilon\le H/4096$,
\begin{equation}
c\frac{SAH^3}{\varepsilon^2}\frac{\kappa(S,m)}{L_A}
\ \le\ N^*(S,A,m,H,\varepsilon)\ \le\
C\frac{SAH^3}{\varepsilon^2}\kappa(S,m)L^3.
\label{eq:main-sandwich}
\end{equation}
If $m$ is at least a sufficiently large constant times $\log(A+1)$, the denominator $L_A$ can be removed from the lower bound, after adjusting the universal constant. The upper bound has a high-probability version for every $\delta\in(0,1)$, with logarithmic dependence on $\delta^{-1}$ specified in Theorems~\ref{thm:chaining} and~\ref{thm:reward-free-simulator}.
\end{theorem}

The proof of this theorem is assembled in Section~\ref{sec:assemble-minimax} after the explicit upper and lower bounds. The logarithmic factors in~\eqref{eq:main-sandwich} depend on $S,A,H,\varepsilon^{-1}$, not on an unspecified polynomial in $m$. In particular, if a tilde hides only logarithms in these other parameters, the joint rate is
\[ 
N^*=\widetilde\Theta\!\left(
\frac{SAH^3}{\varepsilon^2}\min\{S,\log(em)\}
\right).
\]
This statement retains the objective dependence explicitly. When $m$ is exponentially large in $S$, the logarithmic objective cost can itself be of order $S$. For fixed $m$, the bound has the same polynomial dependence as single-objective generative-model planning. The result is minimax optimal up to logarithmic factors, not a constant-factor characterization of every logarithm or of the full confidence dependence.

\textbf{Algorithm.}
The learner samples every state--action pair equally and forms $\widehat P$. In the reward-hull regime, it rounds a query $r_w$ to a finite reward cover and returns an optimal policy for the rounded reward in $\widehat P$. Fixing an order on the cover and on the actions makes both selections deterministic. In the state-saturation regime, it plans directly for $r_w$. The two sufficient sample counts are known before sampling, so the learner chooses the smaller one.

\textbf{Proof structure.}
Section~\ref{sec:chaining} proves the logarithmic-objective upper bound in four steps: construct reward covers, concentrate their value increments, control these increments through a one-sided variance estimate, and compare true and empirical policies. Section~\ref{sec:state-saturation} states the complementary bound for all bounded rewards. Section~\ref{sec:lower-simplex} proves the action-code lower bound, using a common occupancy identity and an adaptive information bound. Section~\ref{sec:assemble-minimax} combines the parameter regimes. Supporting upper and lower arguments appear in the appendices.

At reward scale $\eta$, the cover entropy is of order $\eta^{-2}\log(4SA)\log(em)$. Its square root contributes a factor $\eta^{-1}$ to Bernstein's inequality. The variance comparison supplies a matching factor $H^{3/2}\eta$, together with a term depending on the comparison error. Lemma~\ref{lem:chained-comparison} absorbs that remaining term and sums over the logarithmic number of scales.

\section{The reward-hull upper bound}
\label{sec:chaining}
Let $H=(1-\gamma)^{-1}$, and let
$\mathcal R=\operatorname{conv}\{r_1,\ldots,r_m\}\subset[0,1]^{SA}$.
Rewards are known. All samples below concern transitions. Fix
$\varepsilon\in(0,H]$ and $\delta\in(0,1)$, and define
\begin{align*}
 &J\triangleq\left\lceil\log_2\frac{16H}{\varepsilon}\right\rceil,
\quad \eta_j\triangleq2^{-j}\quad(0\le j\le J),\quad 
 h\triangleq\frac{\varepsilon}{1000HJ},
 \quad \mathcal G\triangleq\{\min\{\ell h,H\}:0\le\ell\le\lceil H/h\rceil\},\\
 &\text{ and }
 \Lambda\triangleq\log\frac{20SAJ|\mathcal G|^2}{\delta}
       +3\log(4SA)\log(em).
\end{align*}
Then $\varepsilon/(32H)<\eta_J\le\varepsilon/(16H)$,
$J\ge4$, and $|\mathcal G|\le1000H^2J/\varepsilon+2$.

\begin{theorem}[Reward-hull chaining upper bound]
\label{thm:chaining}
Draw $n$ independent next states at each state--action pair, form the empirical kernel $\widehat P$, and suppose
\[
 n\ge \frac{2^{20}H^3\Lambda J^2}{\varepsilon^2}.
\]
Then there is a finite $\eta_J$-cover $\mathcal N_J\subset\mathcal R$ in the supremum norm, depending only on the known rewards, with the following property. With probability at least $1-\delta$, simultaneously for every $r\in\mathcal R$, every $q\in\mathcal N_J$ with $\|r-q\|_\infty\le\eta_J$, and every deterministic stationary policy $\widehat\pi$ that is optimal for reward $q$ in the empirical kernel $\widehat P$ (ties broken arbitrarily),
\[
 V_r^*(s)-V_r^{\widehat\pi}(s)\le\varepsilon\qquad\text{for all }s\in\mathcal S.
\]
In particular, the algorithm that chooses any such $q$ and returns any such $\widehat\pi$ is simultaneously $\varepsilon$-optimal for all $r\in\mathcal R$.
The total sample complexity is at most
\[
 SA\left\lceil\frac{2^{20}H^3\Lambda J^2}{\varepsilon^2}\right\rceil.
\]
In particular, the dependence on the number of reward coordinates is logarithmic, and the polynomial dependence is $SAH^3\varepsilon^{-2}$. 
\end{theorem}

\subsection{Finite reward covers}
For $j\ge1$, define
\[
 k_j\triangleq \left\lceil\frac{\log(4SA)}{2\eta_j^2}\right\rceil,
 \qquad
 \mathcal N_j\triangleq \left\{\frac1{k_j}\sum_{\ell=1}^{k_j}r_{i_\ell}:
                (i_1,\ldots,i_{k_j})\in[m]^{k_j}\right\}\subseteq [0,1]^{SA},
\]
and set $\mathcal N_0\triangleq \{0\}$ and $M_j\triangleq |\mathcal N_j|$. Clearly $M_j\le m^{k_j}$. We then claim that  $\mathcal N_j$ is an $\eta_j$-cover of $\mathcal R$. To verify this, consider a fixed reward $r=\sum_iw_ir_i\in \mathcal{R}$. We can sample $k_j$ independent coordinate rewards with index distribution $w$. At each of the $SA$ state-action entries, their average differs from its mean by more than $\eta_j$ with probability at most $2e^{-2k_j\eta_j^2}$. Specifically, Hoeffding's inequality~\cite[Theorem~2]{hoeffding1963probability} applied to a $k_j$-sample average with index distribution $w$ implies 
\[
 \Prob\!\left(\left\|k_j^{-1}\sum_{\ell=1}^{k_j}r_{I_\ell}-r_w\right\|_\infty>\eta_j\right)
 \le2SAe^{-2k_j\eta_j^2},
\]
and a union bound implies  total probability at most $1/2$. Therefore, at least one deterministic average in $\mathcal N_j$ is consequently within $\eta_j$ of $r$ in the supremum norm.

Now for every target $r\in\mathcal N_J$, fix an approximant $q_j(r)\in\mathcal N_j$ with
$\|r-q_j(r)\|_\infty\le\eta_j$ for $1\le j<J$; set $q_0(r)=0$ and $q_J(r)=r$. These choices depend only on the rewards. For every $j\ge1$, we claim 
\[
 \|q_j(r)-q_{j-1}(r)\|_\infty\le3\eta_j.
\]
To prove it, note that for $j=1$, the bound holds as $\|q_1(r)\|_\infty\le1\le3/2$; for $j\ge2$, it uses the fact that $\eta_{j-1}=2\eta_j$ and triangle inequality. Therefore, these approximants form a subset of all adjacent pairs $(q,q')\in\mathcal N_j\times\mathcal N_{j-1}$ with $\|q-q'\|_\infty\le3\eta_j$ (we will use them later in Section~\ref{subsec:increments}).
There are at most $M_jM_{j-1}$ of them, and the union bound at level $j$ runs over these pairs.
Define
\[
 L_j\triangleq \log\frac{20SAJ M_jM_{j-1}|\mathcal G|^2}{\delta},
 \qquad t_j\triangleq L_j/n.
\]
For $j\ge2$, we have $\log(M_jM_{j-1})\le(k_j+k_{j-1})\log m$, with $k_j\le \log(4SA)/(2\eta_j^2)+1$ and $k_{j-1}\le\log(4SA)/(8\eta_j^2)+1$. Because $\eta_j\le1/2$ and $\log(4SA)>1$, the additive constant $2$ is at most $\tfrac12\eta_j^{-2}\log(4SA)$. Hence
\[
 \log(M_jM_{j-1})\le3\eta_j^{-2}\log(4SA)\log(em).
\]
The same bound holds for $j=1$ because $M_0=1$. Since $\eta_j^{-2}\ge1$ and $\log(20SAJ|\mathcal G|^2/\delta)\ge0$, it follows that
\begin{equation}
 L_j\le \Lambda/\eta_j^2.
 \label{eq:log-budget}
\end{equation}
The sample-size condition implies $t_j\le \Lambda/(n\eta_J^2)<1$ for every $j$.

\subsection{Uniform concentration of the value increments}
\label{subsec:increments}
We first present the concentration result used in both upper bounds. The following inequalities are standard in literature.

\begin{lemma}[Bernstein bounds in two variance conventions]
\label{lem:emp-var}
Let $X_1,\ldots,X_n$ be i.i.d. copies of $X$ supported in an interval of length $B$, with $n\ge3$. Write $\mu=\E X$, $\sigma^2=\operatorname{Var}(X)$, $\widehat\mu=n^{-1}\sum_iX_i$, and $\widehat\sigma^2=n^{-1}\sum_i(X_i-\widehat\mu)^2$. For $L>0$, the following bounds have failure probabilities at most $2e^{-L}$ and $4e^{-L}$, respectively:
\begin{align}
 |\widehat\mu-\mu|&\le\sqrt{2\sigma^2 L/n}+2BL/(3n),\label{eq:true-bernstein-mean}\\
 |\widehat\mu-\mu|&\le2\sqrt{L/n}\,\widehat\sigma+4BL/n.\label{eq:empirical-bernstein}
\end{align}
They hold jointly with failure probability at most $6e^{-L}$.
\end{lemma}
The first inequality is Bernstein's inequality. The second follows from Maurer and Pontil~\cite[Theorem~4]{maurer2009empirical}, applied to both tails after translating and rescaling the interval. Their unbiased variance is $n\widehat\sigma^2/(n-1)$; These bounds require $n\ge3$ and $7n/[3(n-1)]\le4$, and every sample-size condition below ensures them.

\begin{lemma}[Simultaneous concentration of adjacent value increments]
\label{lem:increment-concentration}
With probability at least $1-\delta$, the following inequalities hold simultaneously for all $j$, all adjacent reward pairs $(q,q')\in\mathcal N_j\times\mathcal N_{j-1}$ satisfying $\|q-q'\|_\infty\le3\eta_j$, and all state--action pairs:
\begin{align}
 |(\widehat P-P)(v_q(P)-v_{q'}(P))|
 &\le2\sqrt{t_j}\,\sigma_{\widehat P}(v_q(P)-v_{q'}(P))
       +32H\eta_jt_j+10h,\label{eq:true-increment}\\
 |(\widehat P-P)(v_q(\widehat P)-v_{q'}(\widehat P))|
 &\le2\sqrt{t_j}\,\sigma_P(v_q(\widehat P)-v_{q'}(\widehat P))
       +32H\eta_jt_j+10h.\label{eq:emp-increment}
\end{align}
Both inequalities are coordinatewise.
\end{lemma}
\begin{proof}
We will apply an absorbing-state construction to remove the dependence of the empirical-value increment on the sampled row.

For \eqref{eq:true-increment}, first note that the function
$\Delta\triangleq v_q(P)-v_{q'}(P)$ is deterministic and satisfies $\|\Delta\|_\infty\le3H\eta_j$.
Let $Y_1,\ldots,Y_n$ be the samples at $(s,a)$, and apply Lemma~\ref{lem:emp-var}, with $B=6H\eta_j$ and $L=L_j$, to $X_i=\Delta(Y_i)+3H\eta_j\in[0,6H\eta_j]$. Since $P(\cdot\mid s,a)$ and $\widehat P(\cdot\mid s,a)$ are probability vectors, the shift changes neither $(\widehat P-P)\Delta$ nor the standard deviations, and the empirical standard deviation of the $X_i$ is $\sigma_{\widehat P}(\Delta)(s,a)$. Inequality \eqref{eq:empirical-bernstein} therefore implies 
\[
 |(\widehat P-P)\Delta|\le2\sqrt{t_j}\sigma_{\widehat P}(\Delta)+24H\eta_jt_j
\]
with failure probability at most $4e^{-L_j}$ for each row and pair.

For \eqref{eq:emp-increment}, we remove the dependence on the row being sampled, using the absorbing-MDP construction of \cite[Lemmas~8--9]{agarwal2020model}. Fix a state $s$. We modify the empirical MDP by making $s$ absorbing and giving all its actions reward $(1-\gamma)u$, where $u\in[0,H]$. Other states retain reward $q$. Write $v_{s,q,u}$ for its optimal value. This vector is independent of all empirical transition rows at state $s$. It equals $u$ at state $s$. It also satisfies
\begin{align}
 v_{s,q,v_q(\widehat P)(s)}&=v_q(\widehat P),\label{eq:absorption-id}\\
 \|v_{s,q,u}-v_{s,q',u'}\|_\infty
 &\le\max\{|u-u'|,H\|q-q'\|_\infty\}.\label{eq:absorbing-lip}
\end{align}
The first identity is \cite[Lemma~8]{agarwal2020model}. The second extends their boundary Lipschitz bound to two different rewards: let $a$ be the norm on its left. At $s$, the difference is $|u-u'|$; outside $s$, the Bellman inequality bounds it by $\|q-q'\|_\infty+\gamma a$. Hence $a\le\max\{|u-u'|,\|q-q'\|_\infty+\gamma a\}$, which implies the displayed bound.

Now consider $u,u'\in\mathcal G$ with $|u-u'|\le3H\eta_j+2h$ and define the proxy difference
$\Delta_0\triangleq v_{s,q,u}-v_{s,q',u'}$.
Since $h\le H\eta_J/2\le H\eta_j/2$, \eqref{eq:absorbing-lip} implies  $\|\Delta_0\|_\infty\le4H\eta_j$. Conditional on all data outside state $s$, this is a fixed function independent of the $n$ samples at $(s,a)$. Bernstein's inequality consequently implies 
\[
 |(\widehat P-P)\Delta_0|(s,a)
 \le\sqrt{2t_j}\,\sigma_P(\Delta_0)(s,a)+8H\eta_jt_j
\]
with conditional, and hence unconditional, failure probability at most $2e^{-L_j}$. We note the numerical constant $8$ is conservative: the range length is at most $8H\eta_j$.

Then we choose grid points $u,u'$ within $h$ of $v_q(\widehat P)(s)$ and $v_{q'}(\widehat P)(s)$, respectively. The two exact boundary values differ by at most $3H\eta_j$, so this grid pair is among those considered. Equations \eqref{eq:absorption-id}--\eqref{eq:absorbing-lip} give
\[
 \|\Delta-\Delta_0\|_\infty\le2h,
 \text{ where } \Delta\triangleq v_q(\widehat P)-v_{q'}(\widehat P).
\]
Therefore $|(\widehat P-P)(\Delta-\Delta_0)|\le4h$, and
$\sigma_P(\Delta_0)\le\sigma_P(\Delta)+2h$ by the triangle inequality in centered $L_2(P)$. Since $t_j\le1$, we have 
\[
 |(\widehat P-P)\Delta|\le\sqrt{2t_j}\,\sigma_P(\Delta)
       +8H\eta_jt_j+(4+2\sqrt2)h,
\]
which implies \eqref{eq:emp-increment}.
A union bound over at most $SA M_jM_{j-1}$ true-value events, each failing with probability at most $4e^{-L_j}$, and at most
$SA M_jM_{j-1}|\mathcal G|^2$ empirical-proxy events at level $j$, each failing with probability at most $2e^{-L_j}$, has total failure probability at most
\[
 \sum_{j=1}^J 6SA M_jM_{j-1}|\mathcal G|^2e^{-L_j}
 =\frac{3\delta}{10}<\delta.
\]
The union covers every grid pair, so it covers the data-dependent pair chosen above.
\end{proof}

\subsection{A one-sided variance bound for the comparison}
In single-objective RL, \cite{azar2013minimax} utilize the total-variance method to control a value function through its Bellman equation; But here, our objective contains signed differences of optimal values, evaluated under a second kernel. We therefore need the following form.

\begin{lemma}
\label{lem:resolvent-variance}
Let $K$ be an $S\times S$ stochastic matrix, let $R_K\triangleq (I-\gamma K)^{-1}$, and for $u\in\R^S$ define
\[
 \sigma_K(u)(s)\triangleq \sqrt{K(u^2)(s)-(Ku(s))^2}.
\]
Let $c\ge0$ and let $\beta\in\R^S$ satisfy $\beta\ge0$. If $0\le u\le c\mathbf1$ and $b\triangleq  u-\gamma Ku\le\beta$, then
\[
 \gamma^2R_K\sigma_K(u)^2\le2c\,R_K\beta-u^2
 \qquad\text{and}\qquad
 \gamma\|R_K\sigma_K(u)\|_\infty\le\sqrt{2cH\|R_K\beta\|_\infty}.
\]
In particular, if $v\in[0,H]^S$ and $v-\gamma Kv\le2\mathbf1$, then $\gamma\|R_K\sigma_K(v)\|_\infty\le2H^{3/2}$.
\end{lemma}
\begin{proof}
Note that $R_K$ has nonnegative entries \cite{puterman1994} and satisfies $R_K\one=H\one$. In particular, it preserves coordinatewise inequalities.

The relation $\gamma Ku=u-b$ implies  $\gamma^2(Ku)^2=(u-b)^2=u^2-2ub+b^2$. Hence
\begin{align*}
 \gamma^2\sigma_K(u)^2
 =\gamma^2K(u^2)-u^2+2ub-b^2\le\gamma K(u^2)-u^2+2c\beta.
\end{align*}
Specifically, it is from bounding three terms. First, $\gamma^2K(u^2)\le\gamma K(u^2)$, because $K(u^2)\ge0$ and $\gamma^2\le\gamma$. Second, $-b^2\le0$. Third, $2ub\le2u\beta\le2c\beta$: multiplying $b\le\beta$ by $u\ge0$ preserves its direction, and $u\le c\mathbf1$ together with $\beta\ge0$ implies the inequality. 

Apply $R_K$ and use $R_K(\gamma K-I)=-I$:
\[
 \gamma^2R_K\sigma_K(u)^2\le-u^2+2c\,R_K\beta\le2c\,R_K\beta.
\]
This proves the first claim. For each state $s$, the weights $R_K(s,y)$ are nonnegative and sum to $H$, so the Cauchy--Schwarz inequality implies 
\[
 \bigl(R_K\sigma_K(u)(s)\bigr)^2
 \le\Bigl(\sum_yR_K(s,y)\Bigr)\Bigl(\sum_yR_K(s,y)\sigma_K(u)(y)^2\Bigr)
 =H\bigl(R_K\sigma_K(u)^2\bigr)(s).
\]
Multiplying by $\gamma^2$ and using the first claim implies  $\gamma^2(R_K\sigma_K(u)(s))^2\le2cH(R_K\beta)(s)\le2cH\|R_K\beta\|_\infty$. Both $\gamma$ and $R_K\sigma_K(u)(s)$ are nonnegative, so taking square roots and the maximum over $s$ proves the second claim. 

For the last claim, take $u=v$, $c=H$, and $\beta=2\mathbf1$. Then $R_K\beta=2H\mathbf1$ and $\sqrt{2H\cdot H\cdot2H}=2H^{3/2}$ and it completes the proof. 
\end{proof}

\subsection{Chaining the comparison error}
The following lemma will be applied twice, in opposite directions. All target and approximating rewards are the bounded arrays in the reward chain defined above, including the zero reward at level zero. Let $K^0$ and $K^1$ be two transition kernels. Denote the optimal values of the base kernel $K^0$ $v_q\triangleq v_q(K^0)$. For each target $r\in\mathcal N_J$, one can choose a deterministic stationary policy $\pi_r$ that is optimal for $r$ in $K^0$, and define
\[
 R_r\triangleq (I-\gamma K^1_{\pi_r})^{-1},\quad
 f_{r,j}\triangleq v_r-v_{q_j(r)},\quad
 d_{r,j}\triangleq v_{q_j(r)}-v_{q_{j-1}(r)}.
\]
Then we have  $f_{r,J}=0$, $f_{r,0}=v_r$, and
$f_{r,j}=\sum_{k=j+1}^Jd_{r,k}$. Define the finite nonnegative quantity
\[
 E\triangleq \max_{r\in\mathcal N_J,\,0\le j\le J}
 \left\|\gamma R_r\left|(K^1_{\pi_r}-K^0_{\pi_r})f_{r,j}\right|\right\|_\infty.
\]

\begin{lemma}
\label{lem:chained-comparison}
Let $t_j=L_j/n$ be as in Section~\ref{sec:chaining}, so that $t_j\le\Lambda/(n\eta_j^2)$ by \eqref{eq:log-budget}. Suppose that, for all targets $r\in\mathcal N_J$, all levels $1\le j\le J$, and all state--action pairs,
\begin{equation}
 |(K^1-K^0)d_{r,j}|
 \le2\sqrt{t_j}\sigma_{K^1}(d_{r,j})+32H\eta_jt_j+10h.
 \label{eq:generic-concentration}
\end{equation}
Then
\begin{equation}
 E\le36JH^{3/2}\sqrt{\Lambda/n}+\frac{1728H^2\Lambda}{n\eta_J}+20JHh,
 \label{eq:bootstrap-result}
\end{equation}
and $\|v_r(K^0)-V_r^{\pi_r}(K^1)\|_\infty\le E$ for every $r\in\mathcal N_J$.
\end{lemma}

\begin{proof}
We first bound the resolved variance of each tail $f_{r,j}$. We then sum the increment concentration bounds and absorb the resulting term proportional to $\sqrt E$.

\textbf{The Bellman inequality controls every tail.}
Fix a target and level and abbreviate $f=f_{r,j}$ and $\eta=\eta_j$. The reward Lipschitz property implies $|f|\le H\eta\mathbf1$. Optimality of $\pi_r$ for $r$ and Bellman optimality of $v_{q_j(r)}$ imply that 
\[
 v_r=r_{\pi_r}+\gamma K^0_{\pi_r}v_r,
 \qquad v_{q_j(r)}\ge q_j(r)_{\pi_r}+\gamma K^0_{\pi_r}v_{q_j(r)}.
\]
Subtracting shows
$f-\gamma K^0_{\pi_r}f\le\eta\mathbf1$. Moreover, note that the shift $\phi=f+H\eta\mathbf1$ satisfies $0\le\phi\le2H\eta\mathbf1$ and, because $(1-\gamma)H=1$, we have
\[
 \phi-\gamma K^1_{\pi_r}\phi
 =\eta\mathbf1+\bigl(f-\gamma K^0_{\pi_r}f\bigr)-\gamma(K^1_{\pi_r}-K^0_{\pi_r})f
 \le2\eta\mathbf1+\gamma\left|(K^1_{\pi_r}-K^0_{\pi_r})f\right|.
\]
Apply Lemma~\ref{lem:resolvent-variance} with $K=K^1_{\pi_r}$, $u=\phi$, $c=2H\eta$, and $\beta=2\eta\mathbf1+\gamma|(K^1_{\pi_r}-K^0_{\pi_r})f|$. Since $K$ is stochastic, $\sigma_K(\phi)=\sigma_K(f)$. Since $R_r\mathbf1=H\mathbf1$, the definition of $E$ implies  $R_r\beta\le(2H\eta+E)\mathbf1$. Therefore we have that
\begin{equation}
 \gamma\|R_r\sigma_K(f_{r,j})\|_\infty
 \le\sqrt{8H^3\eta_j^2+4H^2\eta_j E}
 \le3H^{3/2}\eta_j+2H\sqrt{\eta_jE}.
 \label{eq:bootstrap-variance}
\end{equation}
This argument also holds at $j=0$, where $q_0=0$, $v_{q_0}=0$, and $\|r\|_\infty\le1=\eta_0$, and trivially at $j=J$, where $f_{r,J}=0$.

\textbf{Increment bounds close the comparison.}
Since $d_{r,j}=f_{r,j-1}-f_{r,j}$, the centered-$L_2$ triangle inequality and $\eta_{j-1}=2\eta_j$ imply that 
\[
 \gamma\|R_r\sigma_K(d_{r,j})\|_\infty
 \le9H^{3/2}\eta_j+5H\sqrt{\eta_jE}.
\]
Use the tail representation of each $f_{r,j}$, the nonnegativity of $R_r$, and \eqref{eq:generic-concentration} at the rows $(s,\pi_r(s))$. These rows give
$|(K^1_{\pi_r}-K^0_{\pi_r})d_{r,j}|\le2\sqrt{t_j}\,\sigma_K(d_{r,j})+(32H\eta_jt_j+10h)\mathbf1$.
Enlarge every tail sum to the full sum; all summands are nonnegative. After taking the maximum defining $E$, we obtain
\begin{align}\label{eq:14}
 E&\le\sum_{j=1}^J\left[
 18H^{3/2}\eta_j\sqrt{t_j}
 +10H\sqrt{\eta_jE}\sqrt{t_j}
 +32H^2\eta_jt_j+10Hh\right]\\
 &\le18JH^{3/2}\sqrt{\Lambda/n}
 +10H\sqrt{\Lambda E/n}\sum_{j=1}^J\eta_j^{-1/2}
 +\frac{32H^2\Lambda}{n}\sum_{j=1}^J\eta_j^{-1}+10JHh\\
 &\le18JH^{3/2}\sqrt{\Lambda/n}
 +40H\sqrt{\frac{\Lambda E}{n\eta_J}}
 +\frac{64H^2\Lambda}{n\eta_J}+10JHh.
\end{align}
The second inequality uses \eqref{eq:log-budget}; the third uses the geometric-sum bounds
$\sum_j\eta_j^{-1/2}\le4\eta_J^{-1/2}$ and
$\sum_j\eta_j^{-1}\le2\eta_J^{-1}$.
Finally, Young's inequality implies
\[
 40H\sqrt{\frac{\Lambda E}{n\eta_J}}
 \le\frac E2+\frac{800H^2\Lambda}{n\eta_J}.
\]
Plugging it in \eqref{eq:14} then proves \eqref{eq:bootstrap-result}.

\textbf{The comparison controls policy evaluation.}
For the last claim, write $V=V_r^{\pi_r}(K^1)$. Subtracting $V=r_{\pi_r}+\gamma K^1_{\pi_r}V$ from $v_r=r_{\pi_r}+\gamma K^0_{\pi_r}v_r$ implies  $(I-\gamma K^1_{\pi_r})(v_r-V)=-\gamma(K^1_{\pi_r}-K^0_{\pi_r})f_{r,0}$, thus we have that 
\[
 v_r-V=-\gamma R_r(K^1_{\pi_r}-K^0_{\pi_r})f_{r,0}.
\]
Since $R_r$ has nonnegative entries, $|R_rx|\le R_r|x|$ for every vector $x$. Hence $\|v_r-V\|_\infty$ is at most the level-zero term in the definition of $E$.
\end{proof}

\subsection{Policy comparison and completion of the proof}
Now we are ready to prove Theorem \ref{thm:chaining}. We will consider the event of Lemma~\ref{lem:increment-concentration}. 

We will apply Lemma~\ref{lem:chained-comparison} first with $K^0=\widehat P$ and $K^1=P$ (inequality \eqref{eq:emp-increment} verifies its hypothesis hence we can apply it). Let $\widehat\pi_r$ be any empirical optimal deterministic stationary policy, and note that since \eqref{eq:emp-increment} concerns values only, the conclusion holds simultaneously for every such choice. The exact policy-evaluation identity is
\[
 v_r(\widehat P)-V_r^{\widehat\pi_r}(P)
 =\gamma(I-\gamma P_{\widehat\pi_r})^{-1}
              (\widehat P_{\widehat\pi_r}-P_{\widehat\pi_r})v_r(\widehat P),
\]
and the last claim of Lemma~\ref{lem:chained-comparison} bounds its norm by $E$. Hence
$\|v_r(\widehat P)-V_r^{\widehat\pi_r}(P)\|_\infty\le\Gamma_n$, where
\[\Gamma_n\triangleq 36JH^{3/2}\sqrt{\Lambda/n}+\frac{1728H^2\Lambda}{n\eta_J}+20JHh.
\]
Apply the same lemma with $K^0=P$ and $K^1=\widehat P$. Inequality \eqref{eq:true-increment} verifies its hypothesis. For a true optimal policy $\pi_r^*$, the lemma implies 
$\|v_r(P)-V_r^{\pi_r^*}(\widehat P)\|_\infty\le\Gamma_n$.

For each target reward and state, empirical optimality implies 
\begin{align*}
 v_r(P)-V_r^{\widehat\pi_r}(P)
=v_r(P)-V_r^{\pi_r^*}(\widehat P)
   +V_r^{\pi_r^*}(\widehat P)-v_r(\widehat P)
   +v_r(\widehat P)-V_r^{\widehat\pi_r}(P)\le 2\Gamma_n\mathbf1.
\end{align*}
The sample-size condition, $\eta_J>\varepsilon/(32H)$, and $J\ge4$ imply
\begin{align*}
 \Gamma_n
\le\left(\frac{36}{1024}
       +\frac{55296}{2^{20}J^2}+\frac1{50}\right)\varepsilon<\frac\varepsilon4.
\end{align*}
Thus every finest-net target has true policy loss at most $\varepsilon/2$.

Finally, for an arbitrary query reward $r\in\mathcal R$, choose $q\in\mathcal N_J$ at distance at most $\eta_J$ and return $\widehat\pi_q$. For any fixed policy, changing the reward by at most $\eta_J$ changes every discounted value by at most $H\eta_J$. The same bound holds for optimal values by \eqref{eq:lipschitz}. Consequently, we have that 
\begin{align}\label{eq:15}
    v_r(P)-V_r^{\widehat\pi_q}(P)
 \le 2H\eta_J\mathbf1+
       v_q(P)-V_q^{\widehat\pi_q}(P)
 \le\left(\frac\varepsilon8+\frac\varepsilon2\right)\mathbf1
 \le\varepsilon\mathbf1.
\end{align}
Namely, the output policy is $\varepsilon$-optimal under $r$. All these events were simultaneous over the finite sets used to answer every query, and the bounds hold for every admissible choice of $q$ and of the empirical optimal policy. \eqref{eq:15} therefore holds simultaneously over the entire reward hull. This hence proves Theorem~\ref{thm:chaining}.

\section{An upper bound independent of the number of objectives}
\label{sec:state-saturation}
From the other perspective, we can apply the similar argument and bound simultaneously to every scalar reward in $[0,1]^{SA}$, and therefore supplies the state-dimension saturation for the reward-hull upper bound.

\begin{theorem}[Uniform planning]
\label{thm:reward-free-simulator}
Let $H=(1-\gamma)^{-1}$, $0<\varepsilon\le H$, and $0<\delta<1$.
Set $h=\varepsilon/(1000H)$, and define
\[
 \mathcal F_h\triangleq 
 \{\min\{jh,H\}:0\le j\le\lceil H/h\rceil\}^{S},
 \qquad
 L_{\mathrm{sat}}\triangleq \log\frac{8SA|\mathcal F_h|}{\delta}.
\]
If $n$ independent transitions are collected at every state--action
pair and
\[
 n\ge2^{14}H^3L_{\mathrm{sat}}/\varepsilon^2,
\]
then, with probability at least $1-\delta$, for every reward
$r\in[0,1]^{SA}$, every empirical optimal deterministic stationary
policy is $\varepsilon$-optimal in the true MDP at every state.
In particular, the total sample is
\[
 O\!\left(\frac{SAH^3}{\varepsilon^2}
 \left[S\log\frac{eH^2}{\varepsilon}
       +\log\frac{SA}{\delta}\right]\right).
\]
\end{theorem}

The proof applies the same one-sided variance comparison to a grid of all value vectors in $[0,H]^S$. It is given in Appendix~\ref{sec:state-proof}. The finite value grid makes the concentration event uniform over every bounded reward and every empirical optimal policy.

\section{Lower bounds}
\label{sec:lower-simplex}
We then study lower bounds in this section. We assume that all rewards are known and bounded by one, and every test uses a simplex preference. The lower bound combines two constructions: a scalar problem supplies the baseline $SAH^3/\varepsilon^2$, and a multi-objective construction which adds a factor proportional to the dimension of an unknown exit distribution. Subset rewards identify that distribution, while a short collection of action rewards isolates the action whose distribution is being tested. 
Our technique, the packing and subset-testing approach, is related to reward-free and finite-known-reward lower bounds~\cite{jin2020reward,chen2026optimal}. Our arguments below imply the discounted simulator construction, a simplex-normalized encoding of the target action pair, and a decoder valid for history-dependent deployment policies.

We first introduce the three ingredients shared by the constructions. The first explains why deployment memory cannot circumvent the lower bound; the other two are standard coding and information inequalities in the form needed here.

\begin{lemma}
\label{lem:rare-exit-occupancy}
Fix $\gamma\in[1/2,1)$, $H=(1-\gamma)^{-1}$, and $p=H^{-1}$.
At a decision state $x$, every action $b$ returns to $x$ with probability
$1-p$ and otherwise exits to an absorbing state $y_j$ according to a
distribution $q_b$. Let the known immediate reward at $x$ be $a_b$,
and let the reward at $y_j$ be a known number $t_j$ under every action.
Set
\[
 D\triangleq \frac{1}{1-\gamma(1-p)}
 =\frac{H}{2-H^{-1}}\ge\frac H2,
 \qquad
 g_b\triangleq  a_b+\gamma\sum_jq_b(j)t_j.
\]
Then, for every randomized history-dependent policy $\pi$ started from $x$,
there is a probability vector $z^\pi$ determined by the policy and the
known rewards, independently of the distributions $q_b$ for a fixed policy, such that
\[
 V^\pi(x)=D\sum_bz_b^\pi g_b,
 \qquad V^*(x)=D\max_b g_b.
\]
\end{lemma}

\begin{proof}
Let $\rho_t(b)$ be the probability of choosing action $b$ at time $t$
conditional on remaining at $x$ through time $t$. Before exit, the
observed state is always $x$, the rewards are known functions of the
chosen actions, and each action has the same survival probability.
Thus $\rho_t$ is a functional of the policy that does not depend on the
unknown exit distributions. Define
\[
 z_b^\pi\triangleq  D^{-1}
 \sum_{t=0}^{\infty}[\gamma(1-p)]^t\rho_t(b).
\]
The geometric series sums to $D$, so $z^\pi$ is a probability vector.
A visit to $x$ contributes the immediate reward $a_b$ and the expected
discounted reward after exit, $\gamma pH\sum_jq_b(j)t_j$.
Since $pH=1$, summing these contributions implies  the value identity.
Every policy therefore implies  a convex combination of the $g_b$, and a
stationary action maximizing $g_b$ attains the upper endpoint.
\end{proof}

\begin{lemma}
\label{lem:balanced-transition-code}
For every even integer $d\ge128$, there is a set
$\mathcal C_d\subset\{-1,1\}^d$ such that each vector has $d/2$
positive entries and
\[
 \log|\mathcal C_d|\ge d/16,
 \qquad
 \operatorname{dist}_{\rm Ham}(\theta,\theta')\ge d/4
 \quad\text{for }\theta\ne\theta'.
\]
\end{lemma}

\begin{proof}
Apply the greedy Gilbert packing argument~\cite{gilbert1952comparison} to the
$\binom d{d/2}\ge2^d/(d+1)$ balanced vectors. A Hamming ball of radius
$\lfloor d/4\rfloor$ in the full cube contains at most
$\exp(dh_2(1/4))$ vectors, where
$h_2(t)=-t\log t-(1-t)\log(1-t)$. Consequently the selected set has
logarithmic size at least
\[
 d\bigl(\log2-h_2(1/4)\bigr)-\log(d+1)\ge d/16.
\]
The last inequality holds for $d\ge128$: the first coefficient exceeds
$0.13$, and $\log(d+1)/d$ decreases on this range and is less than
$0.0675$ at $d=128$.
\end{proof}

\begin{lemma}
\label{lem:exit-information}
Let $d\ge128$ be even, let $0<p\le1$, let $0<\alpha\le1/2$, and write
\[
 u(j)\triangleq \frac1d,
 \qquad q_\theta(j)\triangleq \frac{1+\alpha\theta_j}{d},
 \qquad \theta\in\mathcal C_d.
\]
Consider a family with an integer $R\ge1$ of independently indexed unknown transition
rows. Each such row self-loops with probability $1-p$ and exits with
probability $p$ according to its distribution $q_\theta$; all other rows
are fixed throughout the family. If a learner using at most $N$ adaptive
simulator queries produces an output from which all $R$ codewords can
be decoded with probability at least $3/4$ in every model, then
\[
 N\ge\frac{Rd}{32p\alpha^2}.
\]
\end{lemma}

\begin{proof}
Give the parameter vector $\Theta$ the uniform prior on
$\mathcal C_d^R$. For a fixed parameter vector $\vartheta$, compare the learner's
transcript with the reference model in which every unknown exit
distribution is replaced by $u$. One query to an unknown row has relative
entropy at most
\[
 p\operatorname{KL}(q_{\vartheta_r}\|u)
 \le p\sum_{j=1}^d\frac{(q_{\vartheta_r}(j)-u(j))^2}{u(j)}
 =p\alpha^2.
\]
Here $\operatorname{KL}\le\chi^2$ follows from $\log x\le x-1$.
The adaptive relative-entropy chain rule
\cite[Lemma~15.1]{lattimore2020bandit} implies 
\[
 \operatorname{KL}(\mathbb P_\vartheta^Z\|\mathbb P_0^Z)
 \le Np\alpha^2,
\]
where $Z$ includes the transcript, a model-independent internal random
seed, and the learner's output. Conditional on the past and the seed,
the next query is the same function in both experiments; hence only its
next-state observation contributes to the chain rule. Early stopping
can be padded with queries to a known absorbing state. Apply Fano's
inequality in the reference-distribution form of \cite[Lemma~D.10]{jin2020reward}. The decoder in the hypothesis has
error probability at most $1/4$, so
\[
 Np\alpha^2
 \ge\frac34R\log|\mathcal C_d|-\log2
 \ge\frac12R\log|\mathcal C_d|
 \ge\frac{Rd}{32}.
\]
The middle inequality uses $R\log|\mathcal C_d|\ge d/16\ge8$.
Rearranging proves the claim.
\end{proof}

\Needspace{15\baselineskip}
\subsection{The multi-objective lower bound}

The next construction uses $2^d$ objectives for subset tests and only
$O(\log A)$ additional objectives to select an action pair.

\begin{theorem}[Lower bound with additive objective budget]
\label{thm:simulator-lower-action-code}
Let $A\ge2$, $B\triangleq \lfloor A/2\rfloor$, and
\[
 \ell\triangleq 2\left\lceil\max\{64,8\log B\}\right\rceil.
\]
Let $k\ge1$, let $d\ge128$ be even, and assume
$S\ge k+d$ and $m\ge\ell+2^d$.
For $\gamma\in[1/2,1)$, $H=(1-\gamma)^{-1}$, and
$0<\varepsilon\le H/4096$, there is a family of discounted MDPs with
$S$ states, $A$ actions, and $m$ known reward functions in $[0,1]$ for
which every learner satisfying~\eqref{eq:learning-target} with
probability at least $3/4$ in every model requires
\[
 N\ge2^{-27}\frac{kBdH^3}{\varepsilon^2}.
\]
This holds for adaptive simulator queries and arbitrary randomized
history-dependent output policies.
\end{theorem}

\begin{proof}
We construct the family, show that a successful policy map identifies
all its unknown distributions, and then apply
Lemma~\ref{lem:exit-information}.

\textbf{Constructing the dynamics and the action codes.}
Take decision states $x_1,\ldots,x_k$ and absorbing outcome states
$y_1,\ldots,y_d$; any additional states are absorbing with zero rewards.
At each decision state, pair the first $2B$ actions as
$(b,0),(b,1)$, $b\in[B]$. Every action self-loops with probability
$1-p$, where $p=H^{-1}$. Conditional on exit, action $(b,0)$ uses $u$,
and action $(b,1)$ uses $q_{\theta_{i,b}}$. Here
\[
 \Theta=(\theta_{i,b})\in\mathcal C_d^{kB},
 \qquad \alpha=2048\varepsilon/H\le1/2.
\]
When $A$ is odd, the remaining action duplicates $(1,0)$ in both its
transitions and its known rewards.

To make a preference select one pair, apply
Lemma~\ref{lem:balanced-transition-code} in dimension $\ell$.
Since $\ell\ge128$ and $\ell/16\ge\log B$, it provides $B$ balanced
binary vectors $c_1,\ldots,c_B\in\{0,1\}^{\ell}$, obtained by
replacing signs by zeros and ones. Each vector has $\ell/2$ ones, and
any two differ in at least $\ell/4$ coordinates. Therefore
$v^a\triangleq 2c_a/\ell\in\Delta_\ell$ satisfies
\[
 \langle v^a,c_a\rangle=1,
 \qquad \langle v^a,c_b\rangle\le3/4\quad(b\ne a).
\]
Indeed, equal Hamming weights imply that two distinct vectors share at
most $\ell/2-\ell/8=3\ell/8$ coordinates equal to one.

\textbf{Defining simplex preferences that isolate a pair.}
The first $\ell$ reward coordinates are action rewards:
\[
 r_j(x_i,(b,t))=c_b(j),\qquad r_j(y_z,(b,t))=0.
\]
One further coordinate is supplied for each subset $U\subseteq[d]$:
\[
 r_U(x_i,(b,t))=0,\qquad
 r_U(y_z,(b,t))=\one\{z\in U\}.
\]
All remaining coordinates are zero. These known rewards lie in $[0,1]$
and are identical in every environment. For pair $a$ and subset $U$,
use the simplex preference
\[
 w_{a,U}\triangleq \left(\frac78v^a,\frac18e_U\right),
\]
with zero entries in unused coordinates. Lemma~\ref{lem:rare-exit-occupancy}
expresses the value at $x_i$ through the effective rewards
\[
 g_{b,t}=\frac78\langle v^a,c_b\rangle
       +\frac\gamma8 q_{b,t}(U).
\]
Here $q_{b,0}=u$ and $q_{b,1}=q_{\theta_{i,b}}$ for the fixed decision
state $x_i$.
Both target actions have effective reward at least $7/8$, whereas every
other pair has effective reward at most
\[
 \frac78\frac34+\frac18=\frac{25}{32}<\frac78.
\]
The maximum effective reward is attained within the target pair. If the extra duplicate
exists and $a=1$, it has exactly the value of the target reference action;
it does not change this comparison.

\textbf{Recovering a codeword from the learned policy map.}
Fix a row $(i,a)$ and distinct $\theta,\theta'\in\mathcal C_d$. Set
\[
 U_{\theta,\theta'}\triangleq 
 \{j:\theta_j=1,\ \theta'_j=-1\}.
\]
Balance and separation give $|U_{\theta,\theta'}|\ge d/8$, and
\[
 q_\theta(U_{\theta,\theta'})-u(U_{\theta,\theta'})
 =\frac{\alpha|U_{\theta,\theta'}|}{d},
 \qquad
 q_{\theta'}(U_{\theta,\theta'})-u(U_{\theta,\theta'})
 =-\frac{\alpha|U_{\theta,\theta'}|}{d}.
\]
For preference $w_{a,U_{\theta,\theta'}}$, the optimal target action
therefore changes when the row changes from $\theta$ to $\theta'$.
Their stationary value gap in either model is at least
\[
 D\frac{\gamma\alpha|U_{\theta,\theta'}|}{8d}
 \ge\frac{H\alpha}{256}=8\varepsilon,
\]
using $D\ge H/2$ and $\gamma\ge1/2$.

Let $z_{a,1}$ be the occupancy functional of the output policy from
Lemma~\ref{lem:rare-exit-occupancy}, at this preference and initial state.
If the row is $\theta$, every other action has effective reward at most
that of the target reference action. Consequently its value loss is at
least $8\varepsilon(1-z_{a,1})$, and $\varepsilon$-optimality requires
$z_{a,1}\ge7/8$. If the row is $\theta'$, the value loss is at least
$8\varepsilon z_{a,1}$, so $z_{a,1}\le1/8$. These are restrictions on
the same model-independent functional of the learned policy.

Decode row $(i,a)$ by selecting the codeword $\theta$ whose policy at
$w_{a,U_{\theta,\theta'}}$ has $z_{a,1}\ge7/8$ for every
$\theta'\ne\theta$; use any fixed output if there is no unique candidate.
On the simultaneous success event, the true codeword passes every test.
Every false candidate fails its comparison against the true codeword,
by the preceding negative-gap calculation with the candidates reversed.
Thus the decoder recovers all $kB$ codewords on that event. This is an
information-theoretic decoder; evaluating its occupancy functionals has
no computational role in the argument.

\textbf{Counting simulator observations.}
Lemma~\ref{lem:exit-information} now applies with $R=kB$. Substituting
$p=H^{-1}$ and $\alpha=2048\varepsilon/H$ implies 
\[
 N\ge\frac{kBd}{32p\alpha^2}
 =2^{-27}\frac{kBdH^3}{\varepsilon^2}.
\]
This thus completes the proof.
\end{proof}

Since $\ell=O(\log(A+1))$, choosing $d$ of order
$\min\{S,\log(m-\ell)\}$ and $k=S-d$ implies  the desired
$\min\{S,\log m\}$ factor when $m\ge2\ell$ and that minimum is
sufficiently large. Appendix~\ref{sec:pure-preference-lower} implies  a
related construction that uses only pure preferences and has objective
budget $(A-1)2^d$.

\subsection{The scalar baseline}

For the remaining parameter regimes, we use the following scalar
change-of-measure bound. Its proof, given in
Appendix~\ref{sec:scalar-lower-proof}, reuses
Lemma~\ref{lem:rare-exit-occupancy} to cover history-dependent policies.

\begin{theorem}[Scalar lower bound]
\label{thm:scalar-lower}
Let $S\ge3$, $A\ge2$, $m\ge1$, $\gamma\in[1/2,1)$, and
$H=(1-\gamma)^{-1}$. If $0<\varepsilon\le H/64$, every learner that
uses at most $N$ adaptive generative-model queries and satisfies
\eqref{eq:learning-target} with probability at least $3/4$ on every
instance must satisfy
\[
 N\ge\frac{(S-2)AH^3}{12288\varepsilon^2}.
\]
The conclusion allows arbitrary randomized history-dependent output
policies.
\end{theorem}

\section{Assembling the joint minimax rate}
\label{sec:assemble-minimax}
\begin{proof}[Proof of Theorem~\ref{thm:main}]
We first combine the two upper bounds. Fix $\delta=1/4$. In Theorem~\ref{thm:chaining}, the definitions give
\[
J\le C_1L,\qquad
\log|\mathcal G|\le C_2L,\qquad
\Lambda\le C_3\log(em)L,
\]
for universal constants $C_1,C_2,C_3$. To check these estimates, note that $L=\log(eSAH^2/\varepsilon)$ dominates $1$, $\log(SA)$, $\log(H/\varepsilon)$, and $\log H$ in the stated range $\varepsilon\le H/4096$. Also $\log J\le C L$, so the displayed upper bound on $|\mathcal G|$ implies  the second estimate. Since $\log(em)\ge1$, the terms of $\Lambda$ not containing $m$ can be included in its right-hand side. Therefore the chaining algorithm uses at most
\[
C\frac{SAH^3}{\varepsilon^2}\log(em)L^3
\]
samples. The ceiling adds at most $SA$, which is absorbed by this bound. Theorem~\ref{thm:reward-free-simulator}, whose logarithmic term satisfies $L_{\mathrm{sat}}\le CSL$, implies  at most
\[
C\frac{SAH^3}{\varepsilon^2}SL
\]
samples. Choose the algorithm with the smaller displayed sufficient sample count. Since $L\ge1$, the minimum is at most
$C SAH^3\varepsilon^{-2}\min\{S,\log(em)\}L^3$.

For the lower bound, write $\kappa=\min\{S,\log(em)\}$ and use the action-code length
\[
\ell=2\left\lceil\max\{64,8\log\lfloor A/2\rfloor\}\right\rceil
\le C_4\log(eA).
\]
Suppose first that $m\ge2\ell$ and $\kappa\ge512$. Define
\[
x\triangleq\min\{S/2,\log_2(m-\ell)\},\qquad d\triangleq2\lfloor x/2\rfloor,\qquad k\triangleq S-d.
\]
The assumptions imply $x\ge\kappa/2$: the state term has this property, while $m-\ell\ge m/2$ and the large lower bound on $\log(em)$ imply $\log_2(m/2)\ge\log(em)/2$. Hence $d\ge x-2\ge\kappa/4\ge128$, $d$ is even, and $k\ge S/2$. Moreover $\ell+2^d\le m$. Theorem~\ref{thm:simulator-lower-action-code} applies, and $\lfloor A/2\rfloor\ge A/3$ yields
\[
N^*\ge c_1\frac{SAH^3}{\varepsilon^2}\kappa.
\]

If $\kappa<512$, Theorem~\ref{thm:scalar-lower} implies  the same conclusion after decreasing the universal constant: $S-2\ge S/3$ for $S\ge3$, and the factor $\kappa$ is bounded by 512. The remaining case has $m<2\ell$ and $\kappa\ge512$. Then
\[
\kappa\le\log(em)\le\log(2e\ell)
\le C_5\bigl(1+\log\log(e^e A)\bigr)=C_5L_A.
\]
The scalar lower bound consequently implies 
\[
N^*\ge c_2\frac{SAH^3}{\varepsilon^2}
\ge c_3\frac{SAH^3}{\varepsilon^2}\frac{\kappa}{L_A}.
\]
This proves the universal lower bound. When $m\ge2\ell$, only the first two cases are needed, and the $L_A$ denominator is absent. Since $2\ell\le C\log(A+1)$, the stated sufficient objective-budget condition follows.

Together, they complete the proofs for both upper and lower bounds.
\end{proof}

\section{Discussion}
The objective dependence is determined by the geometry of the known reward hull. A fine cover used in a single union bound would introduce $H^2/\varepsilon^2$ through its entropy. Chaining avoids that loss because finer reward increments have smaller resolved variance. The one-sided Bellman inequality remains valid even though the policy optimal for the target reward need not optimize its approximants.

The logarithmic-objective guarantee applies to empirical planning after reward rounding. Direct empirical planning for every bounded reward has the separate state-saturation guarantee. Both statements assume known bounded reward coordinates, simplex preferences, a stationary discounted kernel, and uniform accuracy at every initial state. The result identifies the joint rate up to logarithmic factors; exact logarithmic and confidence dependence remain separate questions.

\bibliographystyle{ieeetr}
\bibliography{ref}

@string{colt = {Proc. Annual Conference on Learning Theory (CoLT)}}

@article{agarwal2020model,
 author = {Agarwal, Alekh and Kakade, Sham and Yang, Lin F},
 booktitle = {Proc. Annual Conference on Learning Theory (CoLT)},
 journal = {Conference on Learning Theory},
 organization = {PMLR},
 pages = {67--83},
 title = {Model-Based Reinforcement Learning with a Generative Model is Minimax Optimal},
 year = {2020}
}

@article{azar2013minimax,
 author = {Azar, Mohammad Gheshlaghi and Munos, Rémi and Kappen, Hilbert J},
 issn = {1573-0565},
 journal = {Machine learning},
 language = {en},
 month = {June},
 number = {3},
 pages = {325--349},
 publisher = {Springer},
 title = {Minimax PAC bounds on the sample complexity of reinforcement learning with a generative model},
 urldate = {2022-03-08},
 volume = {91},
 year = {2013}
}

@article{gilbert1952comparison,
 author = {Gilbert, Edgar N},
 journal = {The Bell system technical journal},
 number = {3},
 pages = {504--522},
 publisher = {Nokia Bell Labs},
 title = {A comparison of signalling alphabets},
 volume = {31},
 year = {1952}
}

@article{hayes2022practical,
 author = {Hayes, Conor F and Rădulescu, Roxana and Bargiacchi, Eugenio and Källström, Johan and Macfarlane, Matthew and Reymond, Mathieu and Verstraeten, Timothy and Zintgraf, Luisa M and Dazeley, Richard and Heintz, Fredrik and others},
 journal = {Autonomous Agents and Multi-Agent Systems},
 number = {1},
 pages = {26},
 publisher = {Springer},
 title = {A practical guide to multi-objective reinforcement learning and planning},
 volume = {36},
 year = {2022}
}

@inproceedings{jin2020reward,
 author = {Jin, Chi and Krishnamurthy, Akshay and Simchowitz, Max and Yu, Tiancheng},
 booktitle = {International Conference on Machine Learning},
 organization = {PMLR},
 pages = {4870--4879},
 title = {Reward-free exploration for reinforcement learning},
 year = {2020}
}

@article{kaufmann2016complexity,
 author = {Kaufmann, Emilie and Cappé, Olivier and Garivier, Aurélien},
 journal = {The Journal of Machine Learning Research},
 number = {1},
 pages = {1--42},
 publisher = {JMLR. org},
 title = {On the complexity of best-arm identification in multi-armed bandit models},
 volume = {17},
 year = {2016}
}

@book{lattimore2020bandit,
 author = {Lattimore, Tor and Szepesvári, Csaba},
 publisher = {Cambridge University Press},
 title = {Bandit algorithms},
 year = {2020}
}

@inproceedings{li2020breaking,
 author = {Li, Gen and Wei, Yuting and Chi, Yuejie and Gu, Yuantao and Chen, Yuxin},
 booktitle = {Proc. Advances in Neural Information Processing Systems (NeurIPS)},
 editor = {Larochelle, H. and Ranzato, M. and Hadsell, R. and Balcan, M. F. and Lin, H.},
 journal = {arXiv:2005.12900 [cs, math, stat]},
 language = {en},
 month = {December},
 note = {arXiv: 2005.12900},
 pages = {12861--12872},
 publisher = {Curran Associates, Inc.},
 title = {Breaking the sample size barrier in model-based reinforcement learning with a generative model},
 urldate = {2022-01-19},
 volume = {33},
 year = {2020}
}

@article{li2024how,
 author = {Li, Yining and Ju, Peizhong and Shroff, Ness B},
 journal = {arXiv preprint arXiv:2410.15557},
 title = {How to Find the Exact Pareto Front for Multi-Objective MDPs?},
 year = {2024}
}

@inproceedings{lu2023multi,
 author = {Lu, Haoye and Herman, Daniel and Yu, Yaoliang},
 booktitle = {The Eleventh International Conference on Learning Representations},
 title = {Multi-objective reinforcement learning: Convexity, stationarity and pareto optimality},
 year = {2023}
}

@article{maurer2009empirical,
 author = {Maurer, Andreas and Pontil, Massimiliano},
 journal = {arXiv preprint arXiv:0907.3740},
 month = {July},
 note = {arXiv: 0907.3740},
 title = {Empirical Bernstein bounds and sample variance penalization},
 urldate = {2022-03-24},
 year = {2009}
}

@book{puterman1994,
 author = {Puterman, Martin L},
 publisher = {John Wiley \& Sons},
 title = {Markov Decision Processes: Discrete Stochastic Dynamic Programming},
 year = {1994}
}

@article{qiu2024traversing,
 author = {Qiu, Shuang and Zhang, Dake and Yang, Rui and Lyu, Boxiang and Zhang, Tong},
 journal = {arXiv preprint arXiv:2407.17466},
 title = {Traversing pareto optimal policies: Provably efficient multi-objective reinforcement learning},
 year = {2024}
}

@article{rame2023rewarded,
 author = {Rame, Alexandre and Couairon, Guillaume and Dancette, Corentin and Gaya, Jean-Baptiste and Shukor, Mustafa and Soulier, Laure and Cord, Matthieu},
 journal = {Advances in Neural Information Processing Systems},
 pages = {71095--71134},
 title = {Rewarded soups: towards pareto-optimal alignment by interpolating weights fine-tuned on diverse rewards},
 volume = {36},
 year = {2023}
}

@article{roijers2013survey,
 author = {Roijers, Diederik M. and Vamplew, Peter and Whiteson, Shimon and Dazeley, Richard},
 journal = {Journal of Artificial Intelligence Research},
 pages = {67--113},
 title = {A Survey of Multi-Objective Sequential Decision-Making},
 volume = {48},
 year = {2013}
}

@article{wu2021accommodating,
 author = {Wu, Jingfeng and Braverman, Vladimir and Yang, Lin},
 journal = {Advances in Neural Information Processing Systems},
 pages = {13112--13124},
 title = {Accommodating picky customers: Regret bound and exploration complexity for multi-objective reinforcement learning},
 volume = {34},
 year = {2021}
}

@article{yang2019generalized,
 author = {Yang, Runzhe and Sun, Xingyuan and Narasimhan, Karthik},
 booktitle = {Proc. Advances in Neural Information Processing Systems (NeurIPS)},
 journal = {Advances in neural information processing systems},
 title = {A generalized algorithm for multi-objective reinforcement learning and policy adaptation},
 volume = {32},
 year = {2019}
}

@article{zhu2023scaling,
 author = {Zhu, Baiting and Dang, Meihua and Grover, Aditya},
 journal = {arXiv preprint arXiv:2305.00567},
 title = {Scaling pareto-efficient decision making via offline multi-objective rl},
 year = {2023}
}

@article{chen2026optimal,
  title={Optimal Multi-Reward Reinforcement Learning},
  author={Chen, Zijun and Zhang, Zihan},
  journal={arXiv preprint arXiv:2609.36486},
  year={2026}
}

@article{li2026minimax,
  title={Minimax-optimal reward-agnostic exploration in reinforcement learning},
  author={Li, Gen and Yan, Yuling and Chen, Yuxin and Fan, Jianqing},
  journal={Mathematics of Operations Research},
  year={2026},
  publisher={INFORMS}
}

@article{zhou2020provable,
  title={Provable multi-objective reinforcement learning with generative models},
  author={Zhou, Dongruo and Chen, Jiahao and Gu, Quanquan},
  journal={arXiv preprint arXiv:2011.10134},
  year={2020}
}

@article{hoeffding1963probability,
  title={Probability inequalities for sums of bounded random variables},
  author={Hoeffding, Wassily},
  journal={Journal of the American statistical association},
  volume={58},
  number={301},
  pages={13--30},
  year={1963},
  publisher={Taylor \& Francis}
}

@article{mannor2004geometric,
  title={A geometric approach to multi-criterion reinforcement learning},
  author={Mannor, Shie and Shimkin, Nahum},
  journal={Journal of machine learning research},
  volume={5},
  number={Apr},
  pages={325--360},
  year={2004}
}

@article{zhou2024finite,
  title={Finite-time convergence and sample complexity of actor-critic multi-objective reinforcement learning},
  author={Zhou, Tianchen and Hairi, FNU and Yang, Haibo and Liu, Jia and Tong, Tian and Yang, Fan and Momma, Michinari and Gao, Yan},
  journal={arXiv preprint arXiv:2405.03082},
  year={2024}
}

@inproceedings{abels2019dynamic,
  title={Dynamic weights in multi-objective deep reinforcement learning},
  author={Abels, Axel and Roijers, Diederik and Lenaerts, Tom and Now{\'e}, Ann and Steckelmacher, Denis},
  booktitle={International conference on machine learning},
  pages={11--20},
  year={2019},
  organization={PMLR}
}

@article{vaswani2022near,
  title={Near-optimal sample complexity bounds for constrained mdps},
  author={Vaswani, Sharan and Yang, Lin and Szepesv{\'a}ri, Csaba},
  journal={Advances in Neural Information Processing Systems},
  volume={35},
  pages={3110--3122},
  year={2022}
}

@inproceedings{menard2021fast,
  title={Fast active learning for pure exploration in reinforcement learning},
  author={M{\'e}nard, Pierre and Domingues, Omar Darwiche and Jonsson, Anders and Kaufmann, Emilie and Leurent, Edouard and Valko, Michal},
  booktitle={International Conference on Machine Learning},
  pages={7599--7608},
  year={2021},
  organization={PMLR}
}

@article{ridel2026improved,
  title={Improved Bounds for Reward-Agnostic and Reward-Free Exploration},
  author={Ridel, Oran and Cohen, Alon},
  journal={arXiv preprint arXiv:2602.16363},
  year={2026}
}

@article{zhang2020task,
  title={Task-agnostic exploration in reinforcement learning},
  author={Zhang, Xuezhou and Ma, Yuzhe and Singla, Adish},
  journal={Advances in Neural Information Processing Systems},
  volume={33},
  pages={11734--11743},
  year={2020}
}

@inproceedings{
hu2026near,
title={Near-Minimax Multi-Objective {RL} under Predictable Adversarial Preferences and Preference-Free Exploration in Linear {MDP}s},
author={Mingxi Hu and Meiling Yu},
booktitle={Forty-third International Conference on Machine Learning},
year={2026},
url={https://openreview.net/forum?id=4X1XzWFZE3}
}

\appendix
\section{Proof of the state-saturation upper bound}
\label{sec:state-proof}
\begin{proof}
Write $t=L_{\mathrm{sat}}/n$.  The displayed sample condition implies $t\le1$.
For every fixed $f_0\in\mathcal F_h$ and every row, apply
Lemma~\ref{lem:emp-var} with $B=H$ to $X_i=f_0(Y_i)\in[0,H]$, where
$Y_1,\ldots,Y_n$ are the samples at that row. On an event with failure
probability at most $6e^{-L_{\mathrm{sat}}}$, it implies  the following bound; for
$K=\widehat P$ this is \eqref{eq:empirical-bernstein}, and for $K=P$ it
follows from \eqref{eq:true-bernstein-mean} because $\sqrt2\le2$ and
$2/3\le4$:
\[
 |(\widehat P-P)f_0|
 \le2\sqrt t\,\sigma_K(f_0)+4Ht,
 \qquad K\in\{P,\widehat P\}.
\]
A union bound makes these statements simultaneous over rows and grid
points with failure probability at most $\delta$.
For any $f\in[0,H]^S$, choose $f_0\in\mathcal F_h$ with
$\|f-f_0\|_\infty\le h$.  The row error changes by at most $2h$,
and the triangle inequality for centered $L_2(K)$ implies 
$|\sigma_K(f)-\sigma_K(f_0)|\le h$.
Consequently, on the same event,
\begin{equation}
 |(\widehat P-P)f|
 \le2\sqrt t\,\sigma_K(f)+4Ht+4h,
 \quad K\in\{P,\widehat P\},\quad f\in[0,H]^S.
 \label{eq:all-value-concentration}
\end{equation}
All inequalities here are coordinatewise.

We next prove a deterministic evaluation comparison that can be used
in either direction.  Let $(K^0,K^1)$ be either $(P,\widehat P)$ or
$(\widehat P,P)$.  For any reward $r\in[0,1]^{SA}$, let
$f=v_r(K^0)$ and let $\pi$ be a deterministic stationary optimal policy
in the base kernel $K^0$.  Define
\[
 K=K^1_\pi,\quad R=(I-\gamma K)^{-1},\quad
 E=\|\gamma R|(K^1_\pi-K^0_\pi)f|\|_\infty.
\]
The vector $f$ satisfies $0\le f\le H\mathbf1$.  Its Bellman equation
in the base model implies  the one-sided bound
\[
 b\triangleq f-\gamma Kf
 =r_\pi+\gamma(K^0_\pi-K^1_\pi)f
 \le\mathbf1+\gamma|(K^1_\pi-K^0_\pi)f|.
\]
Lemma~\ref{lem:resolvent-variance}, applied with $u=f$, $c=H$, and
$\beta=\mathbf1+\gamma|(K^1_\pi-K^0_\pi)f|$, for which
$R\beta\le(H+E)\mathbf1$ because $R\mathbf1=H\mathbf1$, implies 
\[
 \gamma\|R\sigma_K(f)\|_\infty
 \le\sqrt{2H^3+2H^2E}.
\]
Apply $\gamma R$ to \eqref{eq:all-value-concentration} with the variance
taken under $K^1$ (the true variance when $K^1=P$, the empirical variance
when $K^1=\widehat P$), evaluated at the rows $(s,\pi(s))$. By the row convention, this implies 
\begin{align*}
 E&\le2\sqrt t\sqrt{2H^3+2H^2E}+4H^2t+4Hh\\
 &\le2\sqrt2H^{3/2}\sqrt t
       +2\sqrt2H\sqrt{tE}+4H^2t+4Hh.
\end{align*}
Young's inequality bounds the second term by $E/2+4H^2t$.
Rearranging therefore implies 
\[
 E\le B_n\triangleq 4\sqrt2H^{3/2}\sqrt t+16H^2t+8Hh.
\]
By the definitions of $n$ and $h$, and since $\varepsilon\le H$,
\[
 B_n\le\left(\frac{4\sqrt2}{128}
            +\frac1{1024}+\frac1{125}\right)\varepsilon
 <\frac\varepsilon4.
\]
The exact policy-evaluation identity bounds the difference between
$f=v_r(K^0)$ and the value of $\pi$ under $K^1$ by $E$.

Apply this comparison to a true optimal policy and then to an empirical
optimal policy, using opposite choices of $(K^0,K^1)$.
Empirical optimality makes the middle term in the resulting three-term
value decomposition nonpositive.  The true suboptimality of the
empirical optimal policy is therefore at most $E_{P\to\widehat P}+E_{\widehat P\to P}\le2B_n<\varepsilon/2$. Here the two errors use different policies and resolvents, and each is bounded by $B_n$.
The concentration event covers every vector in $[0,H]^S$, so both
comparisons hold for every reward and every corresponding optimal
policy simultaneously.  Finally,
$|\mathcal F_h|\le(1000H^2/\varepsilon+2)^S$, which implies  the
stated sample-complexity bound.
\end{proof}

\section{The scalar change-of-measure argument}
\label{sec:scalar-lower-proof}

\begin{proof}[Proof of Theorem~\ref{thm:scalar-lower}]
Let $k=S-2$ decision states $x_1,\ldots,x_k$ share two absorbing states
$y_+$ and $y_-$. The first objective rewards one at $y_+$ and zero
elsewhere; all other objectives are zero. We use preference $e_1$.
Every action at a decision state exits with probability $p=H^{-1}$ and
otherwise self-loops. Conditional on exit, it selects $y_+$ with
probability $q_{i,a}$ and $y_-$ otherwise.

The null model has $q_{i,a}=1/2$ for every row. Alternative $(i,a)$
changes only that row to $q_{i,a}=1/2+\alpha$, with
$\alpha=16\varepsilon/H\le1/4$.
Lemma~\ref{lem:rare-exit-occupancy} implies 
\[
 V^\pi(x_i)=D\gamma\sum_a z_{i,a}q_{i,a},
 \qquad V^*(x_i)=D\gamma\max_aq_{i,a}.
\]
Here $z_i$ is a model-independent probability vector determined by the
output policy at $e_1$. In alternative $(i,a)$, the value gap to every
other stationary action is $D\gamma\alpha\ge H\alpha/4=4\varepsilon$.
An $\varepsilon$-optimal output must therefore have $z_{i,a}\ge3/4$.

For fixed $i$, the output events
$\mathcal E_{i,a}=\{z_{i,a}\ge3/4\}$ are disjoint. Thus at least
$A-1\ge A/2$ actions satisfy $\mathbb P_0(\mathcal E_{i,a})\le1/2$,
while the success requirement in their respective alternatives implies 
$\mathbb P_{i,a}(\mathcal E_{i,a})\ge3/4$.
Treat simulator rows as arms. The changed rows have the same positive support, and the number of queries is bounded. The change-of-measure inequality of~\cite[Lemma~1]{kaufmann2016complexity}, followed by binary Pinsker, implies 
\[
 \E_0N_{i,a}\,p\operatorname{kl}(1/2\|1/2+\alpha)
 \ge \operatorname{kl}\bigl(\mathbb P_0(\mathcal E_{i,a}),
                              \mathbb P_{i,a}(\mathcal E_{i,a})\bigr)
 \ge2(3/4-1/2)^2=1/8.
\]
Only queries to row $(i,a)$ differ between the experiments. Their
null-to-alternative divergence satisfies
\[
 p\operatorname{kl}(1/2\|1/2+\alpha)
 =-\frac p2\log(1-4\alpha^2)
 \le\frac{2p\alpha^2}{1-4\alpha^2}\le3p\alpha^2.
\]
The first inequality uses $-\log(1-x)\le x/(1-x)$, and the second
uses $\alpha\le1/4$. The preceding change-of-measure inequality therefore implies
\[
 \E_0N_{i,a}\ge\frac{1}{24p\alpha^2}.
\]
Summing over at least $kA/2$ selected rows and using the maximum query
budget $N$ implies 
\[
 N\ge\frac{kA}{48p\alpha^2}
 =\frac{(S-2)AH^3}{12288\varepsilon^2}.
\]
As in Lemma~\ref{lem:exit-information}, early stopping is handled by
padding with queries to a known absorbing state.
\end{proof}

\section{A lower bound using only pure preferences}
\label{sec:pure-preference-lower}

The following variant uses one objective for each action--subset pair.
It implies  the same logarithmic objective dependence for a fixed number
of actions, and shows that the lower-bound mechanism already appears
when the learner must support only the simplex vertices.

\begin{theorem}[Simulator lower bound for simultaneous preferences]
\label{thm:simulator-lower}
Let $A\ge2$, let $k\ge1$, and let $d\ge128$ be even. Suppose
$S\ge k+d$ and $m\ge(A-1)2^d$.
For $\gamma\in[1/2,1)$, $H=(1-\gamma)^{-1}$, and
$0<\varepsilon\le H/2048$, there is a family of discounted MDPs with
$S$ states, $A$ actions, and $m$ known reward functions in $[0,1]$ for
which every learner satisfying~\eqref{eq:learning-target} with
probability at least $3/4$ in every model requires
\[
 N\ge2^{-25}\frac{k(A-1)dH^3}{\varepsilon^2}.
\]
The conclusion allows adaptive simulator queries and arbitrary
randomized history-dependent output policies.
\end{theorem}

\begin{proof}
Write $B=A-1$, and use the balanced code $\mathcal C_d$ of
Lemma~\ref{lem:balanced-transition-code}. Set
$p=H^{-1}$ and $\alpha=1024\varepsilon/H\le1/2$.
As in the main construction, take decision states $x_1,\ldots,x_k$ and
absorbing outcome states $y_1,\ldots,y_d$. Any additional states are
absorbing with zero rewards. Actions are $0,1,\ldots,B$.
At $x_i$, every action self-loops with probability $1-p$. Conditional
on exit, action $0$ uses $u$, and action $a\in[B]$ uses
$q_{\theta_{i,a}}$, where
$\Theta=(\theta_{i,a})\in\mathcal C_d^{kB}$.

For each $a\in[B]$ and $U\subseteq[d]$, provide the known reward
coordinate
\[
 r^{a,U}(x_i,b)=\frac34\one\{b\in\{0,a\}\},
 \qquad r^{a,U}(y_j,b)=\frac14\one\{j\in U\}.
\]
Pad unused reward coordinates with zero. These rewards lie in $[0,1]$,
and we use only their pure simplex preferences $e_{a,U}$.
Lemma~\ref{lem:rare-exit-occupancy} implies  the effective rewards
\[
 g_b=\frac34\one\{b\in\{0,a\}\}+\frac\gamma4q_b(U).
\]
The target actions $0,a$ have $g_b\ge3/4$, whereas every other action
has $g_b\le1/4$. Thus all other unknown rows are irrelevant to the
optimal comparison.

For distinct codewords $\theta,\theta'$, take
$U_{\theta,\theta'}=\{j:\theta_j=1,\theta'_j=-1\}$.
Its size is at least $d/8$, and $q_\theta(U)-u(U)$ and
$q_{\theta'}(U)-u(U)$ equal respectively
$\alpha|U|/d$ and $-\alpha|U|/d$. The two target stationary actions
exchange optimality, with value gap at least
\[
 D\frac{\gamma\alpha|U_{\theta,\theta'}|}{4d}
 \ge\frac{H\alpha}{128}=8\varepsilon.
\]
The occupancy $z_a$ of the learned policy must therefore satisfy
$z_a\ge7/8$ when the row is $\theta$, and $z_a\le1/8$ when it is
$\theta'$. This follows directly from the convex-combination value
formula: the losses are at least $8\varepsilon(1-z_a)$ and
$8\varepsilon z_a$, respectively.

For each row $(i,a)$, decode a candidate $\theta$ precisely when its
output policies satisfy $z_a\ge7/8$ at every preference
$e_{a,U_{\theta,\theta'}}$, $\theta'\ne\theta$; break a nonunique or
empty candidate set arbitrarily. On the simultaneous success event,
the true codeword passes every comparison, and any false codeword fails
its comparison against the true one. The output thus identifies all
$kB$ codewords. Lemma~\ref{lem:exit-information} implies 
\[
 N\ge\frac{kBd}{32p\alpha^2}
 =2^{-25}\frac{k(A-1)dH^3}{\varepsilon^2}.
\]
\end{proof}

Taking an even $d$ below both $S/2$ and
$\log_2(m/(A-1))$, and setting $k=S-d$, yields
\[
 N\ge c\frac{SAH^3}{\varepsilon^2}
       \min\{S,\log(m/(A-1))\}
\]
when the indicated minimum is sufficiently large. In particular, for
fixed $A$ this is the $\min\{S,\log m\}$ dependence. The main
construction improves its objective budget by using mixed preferences
to encode the target action pair.

\end{document}